\documentclass[preprint,journal]{vgtc}       

\ifpdf
  \pdfoutput=1\relax                   
  \usepackage{graphicx}                
  \DeclareGraphicsExtensions{.pdf,.png,.jpg,.jpeg} 
\else
  \usepackage{graphicx}                
  \DeclareGraphicsExtensions{.eps}     
\fi%

\graphicspath{{figures/}{pictures/}{images/}{./}} 
\usepackage{xspace}
\usepackage{microtype}                 
\PassOptionsToPackage{warn}{textcomp}  
\usepackage{textcomp}                  
\usepackage{mathptmx}                  
\usepackage{times}                     
\usepackage{cite}                      
\usepackage{tabu}                      
\usepackage{booktabs}                  
\usepackage{amsmath}
\usepackage{amssymb}
\usepackage{mathrsfs}
\usepackage[bottom]{footmisc}
\usepackage{graphics}
\usepackage{multirow}
\usepackage{multicol}
\usepackage{color}
\usepackage{transparent}
\usepackage[export]{adjustbox}
\usepackage{appendix}
\usepackage{dblfloatfix}
\usepackage{xcolor}

\preprinttext{To appear in IEEE Transactions on Visualization and Computer Graphics in January 2019.\\ DOI: 10.1109/TVCG.2018.2865027}

\onlineid{1193}

\vgtccategory{Research}
\vgtcpapertype{Application/Design Study}

\newcommand{\toolname}{RetainVis\xspace}
\newcommand{\overview}{Overview\xspace}
\newcommand{\patientsummary}{Patient Summary\xspace}
\newcommand{\patientlist}{Patient List\xspace}
\newcommand{\patientdetail}{Patient Details\xspace}
\newcommand{\patienteditor}{Patient Editor\xspace}
\newcommand{\retainex}{RetainEX\xspace}

\definecolor{OliveGreen}{rgb}{0,0.6,0}

\newcommand{\bc}[1]{\textcolor{black}{#1}}

\usepackage{graphicx}
\usepackage{tikz}

\newcommand*\annotatedFigureBoxCustom[8]{ (#1) rectangle (#2);\node at (#4) [fill=#6,thick,shape=circle,draw=#7,inner sep=2pt,font=\sffamily,text=#8,opacity=0.0] {\textbf{#3}};}
\newcommand*\annotatedFigureBox[4]{\annotatedFigureBoxCustom{#1}{#2}{#3}{#4}{white}{white}{black}{black}}

\newenvironment {annotatedFigure}[1]{\centering\begin{tikzpicture}
\node[anchor=south west,inner sep=0,opacity=0.0] (image) at (0,0) { #1};\begin{scope}[x={(image.south east)},y={(image.north west)}]}{\end{scope}\end{tikzpicture}}

\title{\toolname: Visual Analytics with Interpretable and Interactive\\ Recurrent Neural Networks on Electronic Medical Records}

\author{Bum Chul Kwon, Min-Je Choi, Joanne Taery Kim, Edward Choi,\\ Young Bin Kim, Soonwook Kwon, Jimeng Sun, and Jaegul Choo}

\authorfooter{
\item
 Bum Chul Kwon is with IBM T.J. Watson Research Center. E-mail: bumchul.kwon@us.ibm.com.
\item
 Min-Je Choi, Joanne Taery Kim, and Jaegul Choo, who is the corresponding author, are with Korea University. E-mail: {devnote5676, tengyee, jchoo}@korea.ac.kr.
\item
 Edward Choi and Jimeng Sun are with Georgia Institute of Technology. E-mail: {mp2893, jsun}@gatech.edu.
\item
 Young Bin Kim is with Chung-Ang University. E-mail: ybkim85@cau.ac.kr.
\item
 Soonwook Kwon is with Catholic University of Daegu. E-mail: anatomy3@cu.ac.kr.
}

\shortauthortitle{Kwon \MakeLowercase{\textit{et al.}}: RetainVis: Visual Analytics with Interpretable\\ and Interactive Recurrent Neural Network}

\abstract{
We have recently seen many successful applications of recurrent neural networks (RNNs) on electronic medical records (EMRs), which contain histories of patients' diagnoses, medications, and other various events, in order to predict the current and future states of patients. Despite the strong performance of RNNs, it is often challenging for users to understand why the model makes a particular prediction. Such \textit{black-box} nature of RNNs can impede its wide adoption in clinical practice. Furthermore, we have no established methods to interactively leverage users' domain expertise and prior knowledge as inputs for steering the model. Therefore, our design study aims to provide a visual analytics solution to increase interpretability and interactivity of RNNs via a joint effort of medical experts, artificial intelligence scientists, and visual analytics researchers. Following the iterative design process between the experts, we design, implement, and evaluate a visual analytics tool called \toolname, which couples a newly improved, interpretable, and interactive RNN-based model called \retainex and visualizations for users' exploration of EMR data in the context of prediction tasks. Our study shows the effective use of \toolname for gaining insights into how individual medical codes contribute to making risk predictions, using EMRs of patients with heart failure and cataract symptoms. Our study also demonstrates how we made substantial changes to the state-of-the-art RNN model called RETAIN in order to make use of temporal information and increase interactivity. This study will provide a useful guideline for researchers that aim to design an interpretable and interactive visual analytics tool for RNNs.
} 


\CCScatlist{ 
 \CCScat{K.6.1}{Management of Computing and Information Systems}%
{Project and People Management}{Life Cycle};
 \CCScat{K.7.m}{The Computing Profession}{Miscellaneous}{Ethics}
}

\teaser{
  \centering
  \begin{annotatedFigure}
	{\includegraphics[width=1\linewidth, frame, trim={-2cm 0.5cm -2cm 1.8cm},clip]{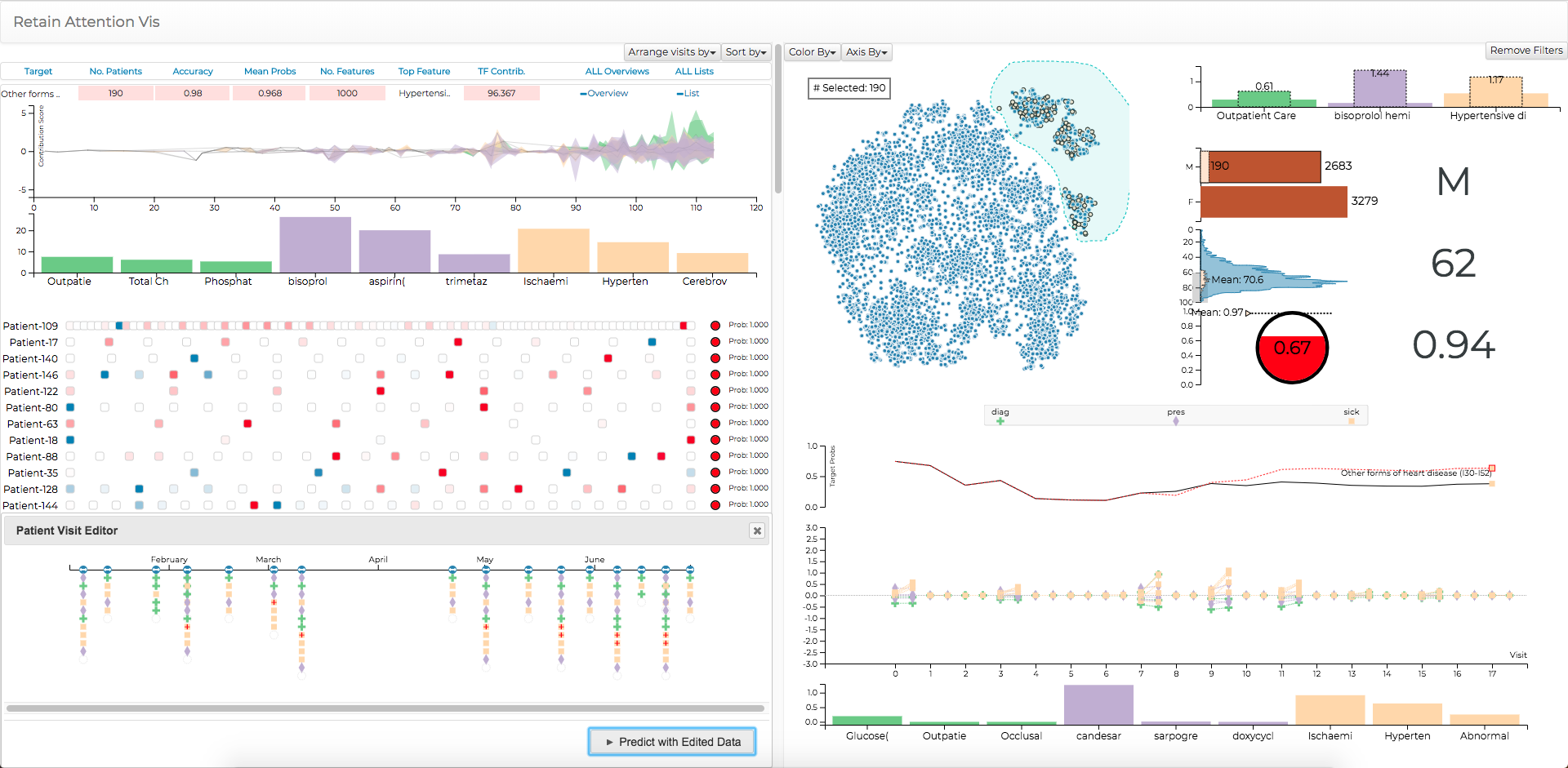}}
	\annotatedFigureBox{0.503,0.4743}{1.0555,0.9472}{A}{0.503,0.9472}
	\annotatedFigureBox{0.03,0.6073}{0.4935,0.9429}{B}{0.03,0.9429}
	\annotatedFigureBox{0.03,0.3328}{0.4935,0.5961}{C}{0.03,0.5961}
	\annotatedFigureBox{0.03,0.0023}{0.4925,0.3306}{D}{0.03,0.3306}
	\annotatedFigureBox{0.503,0.0002}{1.0025,0.4734}{E}{0.503,0.4734}
\end{annotatedFigure}
\caption{A screenshot of \toolname consisting of five areas: (A) \textit{\overview} shows an overview of all patients (left) and an attribute summary view (right) of patients. (B) \textit{\patientsummary} shows the summary of the patient cohort built from (A). (C) \textit{\patientlist} shows individual patients in a row of rectangles. In \patientlist, users can select a patient of interest to view details in (E) \textit{\patientdetail}. Users can open (D) \textit{\patienteditor} to conduct a what-if analysis, and (E) \textit{\patientdetail} shows the updated results.}
	\label{fig:teaser}
}

\vgtcinsertpkg

\begin{document}


\firstsection{Introduction}

\maketitle

In the past decade, we have seen many successful applications of deep learning techniques such as recurrent neural networks (RNNs) on electronic medical records (EMRs), which contain histories of patients' diagnoses, medications, and other events, to predict the current and future states of patients.
This recent movement is related to two key factors.
First, artificial intelligence scientists have been continuously making great advancements in deep learning algorithms and techniques.
Second, although some challenges and concerns (e.g., security and privacy issues) still remain, public and private sectors have started to recognize the needs to make EMRs more accessible to fully leverage the power of deep learning techniques in clinical practice.
These two factors have created a surge of deep learning applications for EMRs, many of which are adopting RNN-based approaches.

Despite the popularity and the ever-increasing performance of RNNs, there exist many challenges to overcome before the full adoption by clinical practice.
A key challenge is for domain experts to understand why the model makes a particular prediction.
The experts also need to be involved in improving the performance of RNNs by providing relevant guidance in order to reduce the risk of costly Type II errors.
Yet, we have no established method to interactively leverage users' domain expertise and prior knowledge as inputs for steering the model.

Thus, our study aims to tackle the problem of intepretability and interactivity by designing a visual analytics solution with an RNN-based model for predictive analysis tasks on EMR data. \bc{Our task is to predict the risk of a patient's future diagnosis in heart failure and cataract, based on information from previous medical visits in our EMR dataset.}
Our design study involved iterative design, assessment, and discussion activities between medical experts, artificial intelligence scientists, and visual analytics researchers.
After we characterized users' tasks, we designed, implemented, and evaluated a visual analytics tool called \toolname with an interactive, interpretable RNN-based model that we name \retainex in order to fulfill the users' needs.

Our study shows the effective use of \toolname for gaining insights into how RNN models EMR data, using real medical records of patients with heart failure and cataract. 
Our study also demonstrates how we made substantial changes to the state-of-the-art RNN model called RETAIN, thereby inventing a new model called \retainex, in order to make use of temporal information and simultaneously increase interactivity and interpretability. 
Various visualizations coupled with the new model allow users to observe the patterns, to test their hypotheses, and to learn interesting stories from patients' medical history.
This study will provide a useful guideline for researchers who aim to design interpretable and interactive visual analytics tools with RNNs.

The following three items summarize our main contributions:

\begin{enumerate}
\vspace{-.1in}
\item We introduce an interpretable, interactive deep learning model, called \retainex, for prediction tasks using EMR data by improving the state-of-the-art model (RETAIN) with additional features for improved interactivity and temporal information.
\vspace{-.1in}
\item We design and develop a visual analytics tool, called \toolname, which tightly integrates the improved deep learning model with the design of visualizations and interactions.
\vspace{-.1in}
\item We conduct both quantitative experiments and a case study with real medical records of patients and discuss the lessons we learned.
\vspace{-.1in}
\end{enumerate}

\vspace{-.1in}
Section~\ref{sec:related} discusses related work from four different perspectives. Section~\ref{sec:users_data_tasks} introduces our target user, data, a prediction variable, and user tasks. Section~\ref{sec:model_description} reviews the backbone model (RETAIN) and our new model (\retainex) with a number of new features for our predictive model. Section~\ref{sec:tool_description} introduces the novel features of our visual analytics system (\toolname). Section~\ref{sec:experiments} shows the quantitative and qualitative experiments we conducted using a real EMR dataset on \retainex and other RNN-based models. Section~\ref{sec:casestudy} shows a case study and Section~\ref{sec:discussions} provides lessons, limitations, and implications learned from our study. Lastly, we conclude this study with future work in Section~\ref{sec:conclusion}.

\vspace{-.05in}
\section{Related Work}
\label{sec:related}
\vspace{-.03in}
This section reviews previous studies using four axes on which our work rests: deep learning applications for EMR data, visualization techniques of black-box models, machine learning platforms that allow user interactivity, and the interpretability of machine learning models.

\textbf{Deep learning for electronic medical records.}
Though the most prevalent use of deep learning techniques in medical domains is to predict diagnosis of a disease, such as breast cancer \cite{araujo17cancer,han17cancer} and brain tumor \cite{havaei17brain,kamnitsas17brain} by training models on medical images, there has also been an increase in deep learning applications for longitudinal EMR data. RNN-based models have been extensively used for tasks such as patient diagnosis \cite{prakash17diag,suo17health}, risk prediction~\cite{choi16doctorai,choi17heart,jin18heart}, representation learning~\cite{choi17gram, choi16med2vec} and patient phenotyping~\cite{che15phenotyping, kale15phenotyping, lipton15phenotyping}, outperforming rule-based and conventional machine learning baselines.

An important issue to consider when designing prediction models using medical data is the interpretability of the model. Medical tasks such as patient diagnosis are performed by clinicians who have sufficient domain knowledge and can explain the reasons of their diagnoses by relating it to past visits of the patient. It is important for machine learning models to incorporate a similar level of interpretability, but many deep learning-based studies in this field fail to address this aspect.

To the best of our knowledge, RETAIN \cite{choi2016retain} is one of the few deep learning models applied to medical data that both harnesses the performance of RNNs and preserves interpretability as to how each data point is related to the output. In RETAIN, it is possible to decompose the output score to the level of individual medical codes that occurred in a patient's previous visits. While there exist other models for EMR data that suggest interpretability such as Dipole \cite{ma2017dipole}, the level of interpretability is limited to each visit, thus failing to provide a complete decomposition of the prediction as RETAIN. For this reason, we use the RETAIN framework for ensuring interpretability of our tool.

\textbf{Visualization of deep learning models.} 
\label{sec:related:visualization}
A major concern in the application of deep learning models is the `black-box' nature. As a result, many approaches have been investigated for visualizing the dynamics in various types of neural networks. Especially for vision applications where convolutional neural networks (CNNs) have enjoyed a great success, many visualization methods such as heatmaps \cite{zintgraf17cnn}, blurring \cite{wang18}, dimensionality reduction \cite{rauber17vis,liu17cnn} of the filters and activation maps obtained during computation and backpropagation, and visualizing the model structure itself \cite{wongsuphasawat18model} were used. This led to a large number of studies dedicated to developing visualization frameworks that help users better understand their networks \cite{kahng18activis,pezzotti18deepeyes,hohman17deep,smilkov17vis, chung2016re}.

Compared to CNNs, RNNs have received less attention in visualization, mainly because of its intertwined structure and its popularity in text data analysis. Though it is possible to visualize the activations of hidden state cells \cite{karpathy15rnn,strobelt18lstmvis, ming17rnnvis}, they do not propose the level of interpretability as in CNNs.
In this aspect, our work makes a substantial contribution in that it aims to provide direct interpretations of the outputs computed using RNNs, supported with a visual analytics tool.

\textbf{Interactive machine learning platforms.} 
A topic of emerging importance in the visual analytics field is to integrate machine learning models with various user interactions \cite{sacha17vis}. Instead of passively observing the outcomes of machine learning models visually, users can make updates to the inputs, outputs, or both, which can further influence the model. This setting enables users to conduct what-if case analyses by adding, editing, or removing data items and then recomputing the outputs. Additionally, a user can instill the model with his/her prior knowledge to correct errors and further improve model performance.

There have been a number of studies to develop tools where users can interact with the results of machine learning tasks such as classification \cite{lin18rclens,ehrenberg16data,heimerl12classifier}, topic modeling \cite{choo13utopian,lee17topic,elassady18topic}, dimensionality reduction~\cite{kwon17axisketcher, cavallo2017exploring} and clustering \cite{lee12cluster,choo15cluster,kwon18cluster}. However, there are only a small number of studies that apply user interaction to tasks requiring deep learning models, such as object segmentation \cite{wang17segment}. To the best of our knowledge, our work is one of the first to apply such user interaction to RNN-based models for medical tasks, supporting direct interaction with the visualized results computed from a deep learning model.

\textbf{Interpretability of deep learning models. }
\bc{The definition of model ``interpretability'' has not been fully established.}
\bc{Model interpretation can be realized in several forms. 
For instance, the weights of a logistic regression model can show what the model has learned. 
In addition, 2-D projection of word embeddings can show how the model interprets each word by showing its distance from others.
Likewise, we can illustrate interpretability using the linear or nonlinear relationship between inputs and outputs, defined by a learned model.
In case of deep learning models, it is difficult to describe the relationship between the input and the output
~\cite{lipton2016mythos}.
In this paper, \retainex aims to achieve this notion of interpretability, similar to weights of logistic regression models.}

\bc{Prior approaches attempted to resolve the interpretability issue in deep learning models.
Visualizing the learned representation is one approach to understand the relationship between the input and the output, as discussed in Section \ref{sec:related:visualization}.
For example, explanation-by-example, such as 2-D projection of word embeddings using t-SNE \cite{maaten2008visualizing} or PCA, can explain model's interpretation of data by showing which words are closely located in the latent space.
An alternative form of interpretation is model-agnostic approaches such as partial dependence plot~\cite{friedman2001greedy}, Shapley values~\cite{shapley1953value} and LIME~\cite{ribeiro2016should}, where they provide some form of explanation as to how a set of features affect the model output.}

\bc{Deep learning models provide interpretable results by showing attention~\cite{luong15attention,bahdanau14attention}. For example, we can imagine a RNN-based model that predicts a sentiment score based on a sentence (i.e., a sequence of words). Given a sentence, the model can compute and assign a score per each word. This score represents importance or contribution that is related to the predicted sentiment score. In our diagnosis risk prediction task, it is natural to think that particular visits of a patient are more important than the rest to predict whether the patient will be diagnosed with heart failure in the future. Therefore, by using the attention mechanism, we can train our model to assign greater weights to more important visits and use these weighted visits for the prediction task. Attention has proven to improve the performance of many sequence-to-sequence based deep learning tasks such as machine translation~\cite{ding18visnmt}, question answering~\cite{xu16vqa}, and speech generation~\cite{wang16speech}.}

\bc{Inspired by the approaches, we aim to increase the interpretability of RNN-based model by  using the attention mechanism as well as visualization methods in the development of \retainex and \toolname.}

\vspace{-.05in}
\section{Users, Data, and Tasks}
\label{sec:users_data_tasks}
\vspace{-.03in}
This section describes target users, input data, and analytic tasks (questions) the users desire to solve.
Based on the description, we review requirements for our model and visualization framework.

\vspace{-.05in}
\subsection{Physicians, Health Professionals, and Researchers}
\vspace{-.03in}
The target users of our visual analytics system include physicians, health professionals, and medical researchers who have access to electronic medical records (EMRs).
They need to answer questions related to diagnosis, prescription, and other medical events.
One of their tasks is to accurately estimate the current and future states of patients.
In addition, they want to investigate the common patterns of patients with the same target outcome (e.g., diabetes).
The experts often want to conduct what-if analysis on patients by testing hypothetical scenarios.

\vspace{-.05in}
\subsection{Data}
\vspace{-.03in}
\label{sec:data}
The dataset used in our visual analytics system, collected between years 2014 and 2015, was provided by the Health Insurance Review and
Assessment Service (HIRA)\cite{kim2014hira},
the national health insurance service in the Republic of Korea. The
HIRA dataset contains the medical information of approximately
51 million Koreans. In particular, the National Patients Sample (HIRA-NPS) dataset consists of information on approximately 1.4 million patients (3\% of all Korean patients) and their related prescriptions. The HIRA-NPS dataset was constructed using age- and gender-stratified
random sampling. The representativeness and validity of this sample
dataset have been confirmed by thorough evaluation against the entire patient population of Korea~\cite{kim2013hira}. The HIRA-NPS contains each patient's encrypted, unique, anonymized identification
(ID) number, medical institution ID number, demographic information,
gender, age, primary and secondary diagnoses, inpatient or outpatient
status, surgical or medical treatment received, prescriptions, and
medical expenses. Each diagnosis is encoded based on the Korean
Standard Classification of Disease, Ninth Revision (KCD-9).
\vspace{-.05in}
\bc{
\subsection{Predicting Diagnosis Risk}
\vspace{-.03in}
Of the various types of tasks in the medical domain where machine learning and deep learning methods can be applied, we chose the task of diagnosis risk prediction. Our medical domain experts were especially interested in the task of predicting whether a patient would later become diagnosed with illnesses such as heart failure using prior visit information. Therefore, we formulated a task setting where we first observe all visits of a patient who has not yet been diagnosed with a target illness (e.g., heart failure), and then predict whether he or she becomes diagnosed with that sickness during a visit in a latter stage, presumably within the next six months. This problem becomes a binary classification task over sequential data, which is a common setting in sequential neural network models as the one we will propose.
}
\vspace{-.05in}
\subsection{User Tasks}
\vspace{-.04in}
\label{sec:tasks}
This section reports our target users' tasks. 
We iteratively identified our target tasks based on weekly discussions among co-authors of this paper, who are experts in visual analytics, deep learning, and medical domains.
We initially generated research questions of our target users' potential interest (led by medical experts), derived visual analytics tasks (led by deep learning and visual analytics experts), and then further evaluated them by closely following a design study methodology~\cite{sedlmair2012design}.
In particular, all leading authors, who were experts in two of the three domains of interest (i.e., deep learning, visual analytics, medical domains), each played the liaison role to fill the gap between domain and technical areas, as Simon et al.~\cite{simon2015bridging} suggests.

The following list shows the main user tasks:

\vspace{-.02in}
\paragraph*{T1:} 
\bc{
\textbf{View Patient Demographics and Medical Records Summary}. Users gain an overview of patients with respect to their demographic characteristics and medical history. The goal of this task is for users to understand the overall distribution of ages, gender, medical history. This enables users to understand the patient groups and to select a subset of them based on their interest.
}
\vspace{-.06in}
\paragraph*{T2:} 
\textbf{Select Interesting Patient Cohorts}. Users test their hypotheses against prior knowledge on specific patient cohorts. In particular, they want to define the cohorts based on various patient attributes.
\vspace{-.06in}
\paragraph*{T3:} 
\textbf{View Summary of Selected Patients} based on visits, medical codes, and prediction scores. Users want to grasp the summary of selected patients. The summary should include the temporal overview of visits, medical codes, and prediction scores.
\vspace{-.06in}
\paragraph*{T4:} 
\bc{
\textbf{Investigate Single Patient History}.
Users investigates a patient's history, especially which visits (i.e., sequence and timing) and medical codes (i.e., event type) contribute to the prediction scores. Users compare contribution scores between medical codes and visits.
}
\vspace{-.06in}
\paragraph*{T5:} 
\textbf{View Contributions of Medical Records for Prediction}. 
Users want to understand why each prediction is made based on patients' visits and medical records. 
In particular, users want to understand the reason by showing the relationship between inputs (patient records) and outputs (prediction scores).
\vspace{-.06in}
\paragraph*{T6:} 
\textbf{Conduct What-If Case Analyses} on individual patients (e.g., add/edit/remove medical code, change visit intervals). Users want to test their hypothetical scenarios on individual patients. For instance, users can check whether the prediction score decreases as they insert a hypothetical visit with a series of treatments.
\vspace{-.06in}
\paragraph*{T7:} 
\textbf{Evaluate and Steer Predictive Model} by viewing summary of prediction scores and providing feedback on contribution scores of visits and medical codes. Users want to check whether the model acts in line with users' prior knowledge. If the model behaves in a undesirable manner, users can provide relevant feedback to the model so that they can improve the model's prediction and interpretation.

By reviewing the tasks, we agreed that a visual analytics system with a recurrent neural network (RNN) model would be a suitable combination to help users accomplish their goals.
In particular, we needed a variant of RNN that can reveal interpretable outcomes.
Thus, we chose the state-of-the-art, interpretable RNN-based model, called RETAIN~\cite{choi2016retain}.
However, RETAIN needed significant improvement in order to fulfill our target users' needs, especially by considering temporal aspects of EMR (i.e., days between visits) and by allowing users to steer the model based on user inputs (\textbf{T3}$-$\textbf{T7}).
In Section~\ref{sec:model_description}, we introduce the improved model called \retainex.
In Section~\ref{sec:tool_description}, we describe the design of our visual analytics tool, called \toolname, and how it fulfills users' needs together with \retainex.

\vspace{-.05in}
\section{Model Description}
\label{sec:model_description}
\vspace{-.03in}
This section describes the structure of our prediction
model, which we name \retainex (RETAIN with extra time dimensions and embedding matrices). We explain the additional features that we incorporated into the original model for greater interactivity, and show how they are capable of fulfilling the user tasks we defined in the previous section.

\vspace{-.05in}
\subsection{Structure of EMR data}
\vspace{-.03in}

A patient's EMR data contain information of a patient's visits over
time. It is usually recorded as a sequence of
medical codes, where each code corresponds to either a doctor's diagnosis of a patient, a treatment or surgery, or a prescribed medicine.
In this sense, we can consider the data of a patient
as a sequence of vectors $\mathbf{x}_{1},\mathbf{x}_{2},\ldots,\mathbf{x}_{T}$,
with $T$ as the total number of visits. For each binary vector $\mathbf{x}_{t}\in\left\{ 0,1\right\} ^{\left|C\right|}$
with $C$ as the number of unique codes, $\mathbf{x}_{t,c}$ is set
to 1 if code $c$ is observed in visit $t$; otherwise set to 0. Note
that each visit may contain more than one code, which results in each
$\mathbf{x}_{t}$ containing multiple values of 1. In this paper,
we focus on using such sequential data on a prediction task, `learning to diagnose\,(L2D)'\,\cite{lipton15diag}, where a model observes the visits of a patient and returns a prediction score indicating the probability of the patient being diagnosed with a target disease in near future.

\vspace{-.05in}
\subsection{\retainex: Interactive temporal attention model}

\begin{figure}[t]
\begin{annotatedFigure}
	{\includegraphics[width=1.0\linewidth]{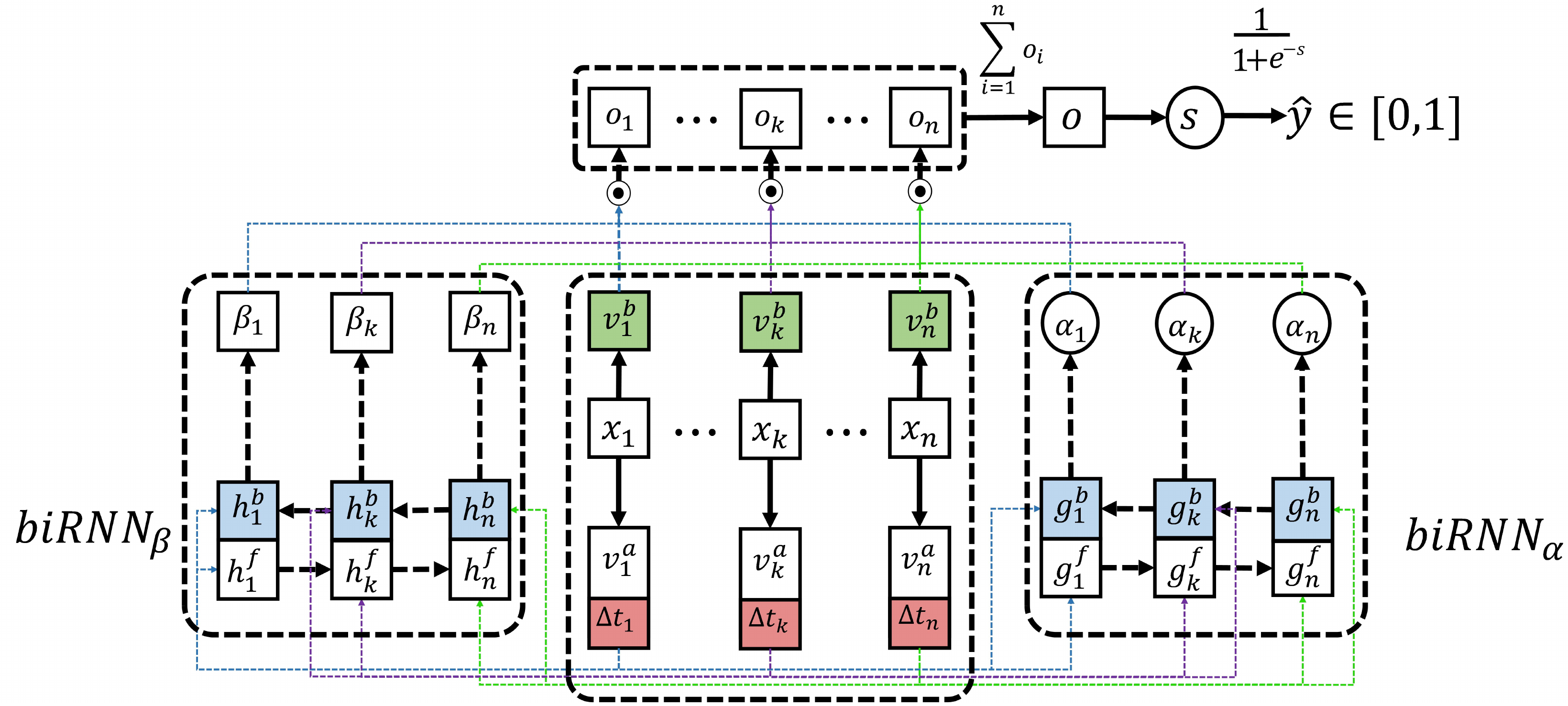}}
	\annotatedFigureBox{0.357,0.4192}{0.6145,0.6092}{A}{0.357,0.6092}
	\annotatedFigureBox{0.116,0.4977}{0.196,0.6077}{B}{0.116,0.6077}
	\annotatedFigureBox{0.644,0.5193}{0.684,0.6093}{C}{0.644,0.6093}
	\annotatedFigureBox{0.362,0.7797}{0.562,0.8997}{D}{0.362,0.8997}
\end{annotatedFigure}
\caption{Overview of \retainex. See Appendix~\ref{app:model_description} for a larger diagram. Our modifications to the original RETAIN model are highlighted by three colors: (i) red~(time information, Section~\ref{sec:time_interval}), (ii) blue~(bidirectional RNNs, Section~\ref{sec:bidirectionality}), and (iii) green~(an additional embedding matrix, Section~\ref{sec:model_interpret}). (A) Using separate embedding matrices,
a binary vector $\mathbf{x}_{t}$ is represented as embedding vectors $\mathbf{v^{\mathit{a}}_{t}}$
and $\mathbf{v^{\mathit{b}}_{t}}$,
with time interval information appended to the former. (B,~C) $\mathbf{v^{\mathit{a}}_{t}}$ is fed into two biRNNs to obtain scalar $\alpha$ and vector $\beta$ attention weights (D) $\alpha$, $\boldsymbol{\beta}$ and $\mathbf{v^{\mathit{b}}}$
are multiplied over all timesteps, then summed to form a single
vector $\boldsymbol{o}$, which is linearly and nonlinearly transformed to a probability score $\hat{y}$.}
\label{fig:overview}
\vspace{-.2in}
\end{figure}

\vspace{-.05in}

As Figure~\ref{fig:overview}~(A) shows, our model takes the patient visit sequence
as $C$-dimensional vectors $\mathbf{x}_{1},\mathbf{x}_{2},\ldots,\mathbf{x}_{T}$
along with the time intervals between each visit, $\Delta t_{1},\Delta t_{2},\ldots,\Delta t_{T}$.
Our model uses two embedding matrices 
$\mathbf{W}_{emb}^{a}\in\mathbb{R^{\mathit{m\times C}}}$
and $\mathbf{W}_{emb}^{b}\in\mathbb{R^{\mathit{m\times C}}}$ to convert
the binary vectors into continuous vectors. We obtain a representation vector for each visit as $\mathbf{v}_{t}^{a}=\mathbf{W}_{emb}^{a}\mathbf{x}_{t}$.
The vectors
$\mathbf{v}_{1}^{b},\ldots,\mathbf{v}_{T}^{b}$ are obtained likewise.
As each visit is associated with a time interval, we compute and use the three
different time values (see details in Section~\ref{sec:time_interval}) 
per vector $\mathbf{v}_{t}^{a}$ .

Figures~\ref{fig:overview}~(B) and (C) represent the bidirectional RNNs that take in the
time-attached visit representations and return attention values (e.g., contribution scores) of different scales. \bc{We follow the original RETAIN settings and compute two attention types, $\mathbf{\alpha}$ and $\mathbf{\beta}$. For the $t$th visit, $\mathbf{\alpha}_{t}$ is a single value that represents the importance of the particular visit. Meanwhile, $\mathbf{\beta}_{t}$ is a $m$-dimensional vector that represents the importance of each medical code within a particular visit. As $\mathbf{\alpha}$ and $\mathbf{\beta}$ are two separate attention types, our model uses two RNN modules. We included the details of how RNNs are computed in Appendix~\ref{app:rnn}.}

We also modified the unidirectional RNN modules to bidirectional ones, as discussed in Section~\ref{sec:bidirectionality}. For each $\mathbf{v}_{t}^{a}$, $\mathrm{biRNN_{\mathit{\alpha}}}$
computes the forward and backward hidden states, $\mathbf{g}_{t}^{f}$
and $\mathbf{g}_{t}^{b}$, which are concatenated as a single $2m$-dimensional
vector. We use a parameter 
$\mathbf{w}_{\alpha}\in\mathbb{R^{\mathit{2m}}}$
to compute a scalar value for each timestep as $e_{t}=\mathbf{w}_{\alpha}\left[\mathbf{g}_{t}^{f};\mathbf{g}_{t}^{b}\right]$.
Then, we apply the softmax function on all scalar values $e_1, \ldots, e_T$ to obtain
$\alpha_{1},\alpha_{2},\ldots,\alpha_{T}$, a distribution of attention
values that sum to one. Similarly, the concatenated hidden state vectors
generated using $\mathrm{biRNN_{\mathit{\beta}}}$ are multiplied
by $\mathbf{W}_{\beta}\in\mathbb{R^{\mathit{\mathrm{m}\times2m}}}$
and return an $m$-dimensional vector 
$\mathbf{\beta}_{t}$
for the $t$-th timestep as $\mathbf{\beta}_{t}=\mathbf{W}_{\alpha}\left[\mathbf{h}_{t}^{f};\mathbf{h}_{t}^{b}\right]$
.

Once we obtain both alpha and beta values, we multiply these values
with the other set of embedding vectors, $\mathbf{v}_{1}^{b},\ldots,\mathbf{v}_{T}^{b}$ as in Figure~\ref{fig:overview}~(D),
and add up the values to obtain the context vector $\mathbf{o}$ 
 with $\odot$ indicating elementwise multiplication of two vectors.
Lastly, we compute the \textit{final contribution score}, $s=\mathbf{w}^{T}_{out}\mathbf{o}$. \bc{This scalar value is transformed to compute a prediction value $\hat{y}=\frac{1}{1+e^{-s}}$, ranging between 0
and 1 where $\mathbf{w}_{out}\in\mathbb{R^{\mathit{m}}}$. The predicted value indicates the diagnosis risk of a patient, with a value closer to 1 indicating a higher risk.}
We train our model by optimizing all parameters to minimize
the cross-entropy loss, $\mathfrak{\mathcal{\mathscr{L}}}=-\frac{1}{N}\sum_{i=1}^{N}y_{i}\log\left(\hat{y_{i}}\right)+(1-y_{i})
\log(1-\hat{y_{i}})$,
 with $y_{i}$ as the target value for the $i$-th patient
among all patients ($N$).

\vspace{-.05in}
\subsection{Bidirectionality}
\label{sec:bidirectionality}
\vspace{-.03in}
Compared to traditional RNNs, which process the input sequence in one direction from the beginning to the end, bidirectional RNNs\,(biRNNs) introduce another set of hidden state vectors computed in a reverse order from the end to the beginning.

In EMR-based diagnosis, clinicians can observe a patient's history in a chronological
order to see how the patient's status progresses over time and also
trace backward in a reverse order from the end to identify possible cues that may strengthen or weaken
their confidence of the patient's current state. 
While the original RETAIN model uses unidirectional RNNs in a reverse direction, we formulate a more intuitive and accurate prediction model by processing the input data with biRNNs. The structure of a bidirectional RNN is discussed in Appendix~\ref{app:bidirectionality}.

\vspace{-.05in}
\subsection{Data with non-uniform time intervals}
\vspace{-.03in}
\label{sec:time_interval}
Though general RNNs do not consider time intervals between visits, the temporal aspect is a key to the disease diagnosis. 
For instance, a burst of the same events over a short time period may forebode the manifestation of a serious illness, while a long hibernation between events may indicate that they may not be influential for diagnosis.

To harness temporal information, we incorporate visit dates as an additional feature to the input vectors of our RNN model.
Given a sequence of $T$ timestamps $t_{1},t_{2},\ldots,t_{T}$, we
obtain $T$ interval values $\Delta t_{1},\Delta t_{2},\ldots,\Delta t_{T}$
with $\Delta t_{i}=t_{i}-t_{i-1}$. We assume that the first visit is unaffected by time constraints by fixing $\Delta t_{1}$ to 1.
For each $\Delta t_{i}$ we calculate different time representations of a single interval, which are (1) $\Delta t_{i}$ (the time interval itself), (2)  $1/\Delta t_{i}$ (its reciprocal value), and (3) $1/\log\left(e+\Delta t_{i}\right)$ (an exponentially decaying value). The first representation was introduced in~\cite{choi2016retain} where time interval information was incorporated to RETAIN, and the latter two were proposed by~\cite{baytas17timelstm}. These three values are concatenated to the input vectors of each step, to enrich the information for our model.
We added the three representations of time intervals because our model can learn to use multiple types of time information and their contributions to prediction results. 
Experimental results in Section~\ref{sec:experiments} show that the addition of time information significantly improves predictive performance.




\vspace{-.05in}
\subsection{Understanding the interpretability of \retainex}
\vspace{-.03in}
\label{sec:model_interpret}

In this section, we show how we achieve interpretability by computing contribution scores using the attention mechanism in \retainex.


\vspace{-.05in}
\paragraph*{\textbf{T4\&T5: Understanding how predictions are made}.}
\retainex achieves its transparency by multiplying the RNN-generated attention weights $\alpha_{t}$s and $\mathbf{\boldsymbol{\beta}_{\mathrm{t}}}$s to the visit vectors $\mathbf{v}_{t}$ to obtain the context vector $\mathbf{o}$, which is used, instead of the RNN hidden state vectors, to make predictions. 
Each input vector $\mathbf{x}_{t}$ has a linear relationship
with the final contribution score $s$.
Thus, we can formulate an equation
that measures the contribution score of the code $c$ at timestep
$t$ to $s$ by reformulating the aforementioned equations as $s_{t,c}=\alpha_{t}\mathbf{w}_{out}\left(\mathbf{\mathit{\mathbf{W}_{emb}^{b}\left[\mathrm{:,\mathit{c}}\right]\odot\boldsymbol{\beta}_{t}}}\right)$,
where $\mathbf{W}_{emb}^{b}$[:,$c$] is the $c$-th column of $\mathbf{W}_{emb}^{b}$.

In our model, we provide two levels of interpretability: visit- and
code-level. The code-level contribution score is the contribution
score of code $c$ at timestep $t$ as described in the above equation.
We can derive a visit-level contribution score $s_{t}$ 
by aggregating contribution scores of codes for each visit 
as $s_{t}=\sum_{c\in\mathbf{x}_{t}}s_{t,c}$.

\paragraph*{\textbf{T3: Summary of selected patients}.}
It is possible to create a vectorized representation of each patient
using the learned contribution scores. 
We assign a 1400-dimensional zero
vector $\mathbf{S}$ to each patient, compute all individual contribution
scores for all codes in every visit that a patient had, and add the
contribution score of each one code (e.g., $s_{t,c}$) to the corresponding
row of $\mathbf{S}$, i.e., $\mathbf{S\left[\mathit{c}\right]}$. 
The dimension size of 1,400 is due to our preprocessed dataset containing 500 treatment codes, 268 diagnosis codes, and 632 prescription codes (see details in Section~\ref{sec:experimentalsetup}). 
The resulting $\mathbf{S}$ can be seen as a patient embedding whose sum of elements and direction each indicate the predicted diagnosis risk and distribution
of input features. 
We later use these vectors to create the scatter plot view in fig.~\ref{fig:teaser}~(A) for exploratory analyses.

\vspace{-.1in}
\subsection{Interactions featured in \retainex}
\vspace{-.03in}
We increase the interpretability of the contribution score by allowing users to experiment with how outputs change according to input changes in an interactive manner.
We provide three ways to edit inputs: adding or removing
input codes, modifying visit periods, 
and modifying the contribution scores of individual codes.

\vspace{-.03in}
\paragraph*{\textbf{T6: Conducting what-if case analysis}.}

Adding or removing codes only requires a simple modification to $\mathbf{x}_{t}$ by changing an element to 1 or 0 with all other input vectors fixed, and then feeding all inputs into the model again for recomputation. 
For modification of time intervals, once a time interval $\Delta t_{i}$ changes,
the corresponding three time values are also updated, and put into the model for recomputation. 
It is a simple and effective scheme for incorporating temporal information
into the model, which also guarantees improved performance.

\vspace{-.03in}
\paragraph*{\textbf{T7: Evaluating and steering predictive model}.}

This interaction allows users to provide feedback onto contribution scores of individual visits or codes based on their domain expertise.
While the earlier two types of interaction are straightforward in that we modify the inputs and insert them into the model to obtain new results, the third type of interaction is more
complex since the model now has to update the learned parameters according to the user's
actions so that it can place more weight to the specified inputs without
harming the overall distribution of attentions assigned to different visits
and codes.

In the original RETAIN, the visit embeddings $\mathbf{v}_{1},\ldots,\mathbf{v}_{T}$,
which are obtained from the binary vectors $\mathbf{x}_{1},,\ldots,\mathbf{x}_{T}$
and the embedding matrix $\mathbf{W}_{emb}$, are used to both (1)~compute
alpha and beta weights, and (2)\,compute the final outputs by multiplying
itself with the alpha and beta values and then summing up
across all timesteps to form a single context vector $\mathbf{o}$.
As we want to change the contributions of specific code(s) at a particular visit
without changing the alpha and beta attentions at other visits, we
formulate an optimization problem where we minimize $\mathit{\mathcal{\mathscr{\mathcal{L}_{\mathit{retrain}}}}}=e^{-s_{pos}+s_{neg}}$
with $s_{pos}$ and $s_{neg}$ being the sums of user-selected contribution
scores $s_{t,c}$ to either increase or decrease. The
retraining process thus is equivalent to performing a number
of gradient descent operations to the parameters, which we restrict
to $\mathbf{W}_{emb}$.

\bc{
Our loss function has to take a monotonically increasing form while maintaining a positive value. We also have to train our model to increase/decrease a number of contribution scores at the same time. As can be seen in the paper, minimizing $\mathcal{L}_{\mathit{retrain}}$ is equivalent to maximizing positive contributions while minimizing negative contributions.}

\bc{Though nonlinear functions such as sigmoid and tanh functions can be used for computing the loss, we chose the exponential function. When having to reduce negative contribution scores, we discovered that the input value $\mathit{-s_{pos}+s_{neg}}$ is often a real number larger than 2. While this would result in saturated gradients close to 0 for the aforementioned nonlinear functions, the exponential function would get a high gradient value from this calculation. This value combined with an adequate learning rate can optimize the parameters of the embedding matrix.
}

To preserve the overall attention distribution while changing the weights
of specific medical codes, the embeddings used to calculate alpha and
beta values need to be separated from the embeddings used for retraining.
Thus, we apply relaxation to our model by introducing two embedding matrices $\mathbf{W^{\mathit{a}}_\mathit{{emb}}}$ and $\mathbf{W^{\mathit{b}}_\mathit{{emb}}}$, subsequently producing two sets of visit embeddings
$\mathbf{v^{\mathit{a}}}_{1},\ldots,\mathbf{v^{\mathit{a}}}_{T}$
and $\mathbf{v^{\mathit{b}}}_{1},\ldots,\mathbf{v}_{T}^{b}$. 
The first set is used to compute the alpha and beta attention weights, while the weights 
are multiplied to the second set for the final outputs. 
Due to the relaxation, we can control the
influence of individual codes without altering the overall distribution of attention
with respect to $\mathbf{W^{\mathit{b}}_\mathit{{emb}}}$.

\bc{While our retraining scheme helps improve the performance of our model, we also want to maintain the real-time interactivity. 
The retraining process is completed in real-time as the only parameters that are actually modified during the process are the weights from $\mathbf{W^{\mathit{b}}_\mathit{{emb}}}$. From a deep learning prospective, backpropagating the weights of a single embedding matrix from a single data sample can be done in milliseconds. 
Throughout experiments, we discovered that retraining was the most effective when optimized for around 10 to 20 iterations with a learning rate of 0.01. 
It took an average of 0.015 seconds to retrain the model according to the data of a patient with 20 visits where 5 codes were modified. 
Due to the simplicity of the scheme, the model was updated within a second after 20 iterations of retraining.
}


\vspace{-.05in}
\section{\toolname: Visual Analytics with \retainex}
\label{sec:tool_description}
\vspace{-.03in}

This section introduces visualization and interaction features that integrate \retainex and describe how the design fulfills user tasks.


\vspace{-.05in}
\subsection{\overview}
\vspace{-.03in}

\begin{figure}[t]
\centering
\begin{annotatedFigure}
	{\includegraphics[width=1.0\linewidth, frame, trim={0 0.1cm 0 0.1cm},clip]{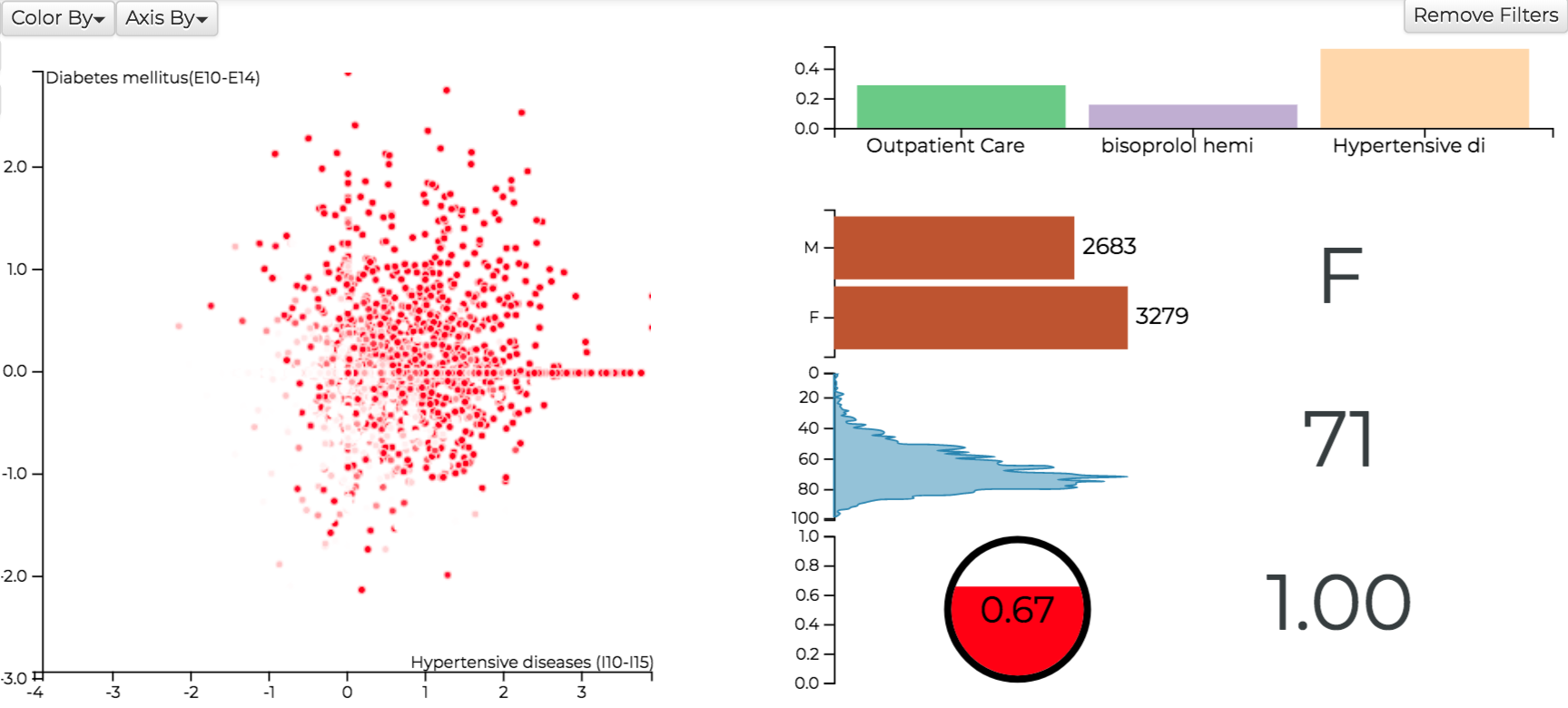}}
	\annotatedFigureBox{0.05,0.03}{0.4457,0.915}{A}{0.05,0.915}
	\annotatedFigureBox{0.497,0.7697}{1.035,0.915}{B}{0.497,0.915}
	\annotatedFigureBox{0.497,0.0059}{0.763,0.7173}{C}{0.497,0.7173}
\end{annotatedFigure}
\caption{\emph{\overview} shows all patients in (A) a scatter plot; (B) A bar chart shows the top three contributors and their mean scores; (C) Three charts (i.e., bar chart, area chart, and circle chart) show the mean and distribution of gender, age, and predicted diagnosis risks, respectively.}
\label{fig:retain_overview_view}
\vspace{-.2in}
\end{figure}

\vspace{-.02in}
\overview aims to provide a summary of patients with respect to their medical codes, contribution scores, and predicted diagnosis risks (\textbf{T1}).
To derive this overview, we use $\mathbf{S}$ (see Section~\ref{sec:model_interpret} for details), which is a list of all patients vectorized with contribution scores of medical codes.
We ran \textit{t-SNE} on $\mathbf{S}$ to derive two dimensional vector list in order to projects patients on 2-D space.
Figure~\ref{fig:retain_overview_view}~(A) shows that \overview depicts patients' differences by the distance between points. 

In \overview, users can choose to map between patient attributes (e.g., age, gender, contribution scores) and two graphical variables: color and position (axes), \bc{as Figure~\ref{fig:retain_overview_view}~(A) shows}.
For instance, users can map predicted diagnosis risks to a linear color scale of white to red (0 to 1) as shown in Figure~\ref{fig:retain_overview_view}~(A).
Users can also show male and female patients in different colors.
Then, users can also switch axes by choosing two out of any attributes.
Figure~\ref{fig:retain_overview_view}~(A) shows that the user chose two comorbidities of heart failure patients, namely hypertensive diseases (x) and diabetes mellitus (y).
The chart shows the model's overall high predicted risks around the region except for lower left corners--which indicates patients with low contribution scores of both hypertension and diabetes.
From the view, we can hypothesize that predicting patients without strong contributions of any of these comorbidities will be difficult for the model.

The right side of \overview shows four charts: code bar chart, gender bar chart, age area chart, and prediction circle chart, from top to bottom.
The four charts mainly summarize patients by their attributes.
To avoid overplotting, we only show the top three highest contributors in code bar chart.
\bc{This approach has limitations, where it only shows three measures at a time. We could have fit more bars by showing narrower, horizontal bars. It is certainly a direction that future designers can implement. In this particular implementation, we wanted to maintain the consistency between other bar charts in other views, where the bar height was consistently used as an indicator for contribution scores.
}
The contribution scores were computed by patient-wise mean of the corresponding codes in score vectors of patients ($\mathbf{S}$).
Users can see the distributions of age and gender in gender bar chart and age area chart, respectively.
Prediction circle chart shows the mean predicted diagnosis risk as the gauge filled in the circle.
This particular icon is consistently used to show individual patient's predicted diagnosis risk in \patientlist as well.
The bottom right corner of \overview shows gender, age, and predicted risk of a selected/highlighted patient.

The five charts in \overview not only serves as summary of patients but also acts as custom cohort builders (\textbf{T2}).
Using coordinated interaction between the five charts, users can define customized patient groups by setting filters on each view.
\bc{The scatter plot view initially shows 2D projection of patients using a nonlinear manifold learning technique, t-SNE.}
In scatter plot, users can draw a polygonal area with a lasso drawing tool.
\bc{
Our interaction approach, namely Lasso-selection in 2D projection derived from multidimensional data, has limitations. The 2D projection cannot provide the most faithful distances between points. Users may not be able to correctly express their regions of interests due to such information loss. Many prior techniques attempt to expose such artifacts of dimension reduction (e.g., CheckViz~\cite{lespinats_checkviz_2013}) and to resolve the issues by providing some guidance~\cite{heulot_proxilens_2013, aupetit_multidimensional_2014}. Alternatively, we can provide visual summaries of axes of nonlinear projection (e.g.,  AxiSketcher~\cite{kwon17axisketcher}). 
In order to provide more options for scatter plot axes, we allowed users to select and switch axes to other measures, such as prediction certainty, contribution scores of medical codes, and demographic information (e.g., age).
The 2D projection view may also suffer the clutter issue, which can be minimized by showing only representative points based on clustering as shown in Clustervision~\cite{kwon18cluster}.
}
Once users complete drawing, the points surrounded by the drawn region are highlighted.
In addition, other views quickly show a summary of the highlighted points: 1) dotted bars for the mean values of selected patients in code bar chart; 2) distributions of selected patients in yellow bars and yellow areas in gender bar chart and age area chart, respectively; and 3) mean predicted diagnosis risk as a dotted horizontal line in prediction circle chart.
Users can also set filters in other views: by clicking bars or brushing axes.
Thus, \overview highlights patients that satisfy all conditions set by users.
For instance, users can select a small cohort of six patients by drawing an area representing positive contributions from both ischaemic heart diseases (x) and pulmonary heart diseases (y) as well as choosing the age group between 60 and 80.


\vspace{-.05in}
\subsection{\patientsummary}
\label{sec:patientsummary}
\vspace{-.03in}

\begin{figure}[t]
\centering
\includegraphics[width=1\linewidth, frame, trim={0 0.1cm 0 0.1cm},clip]{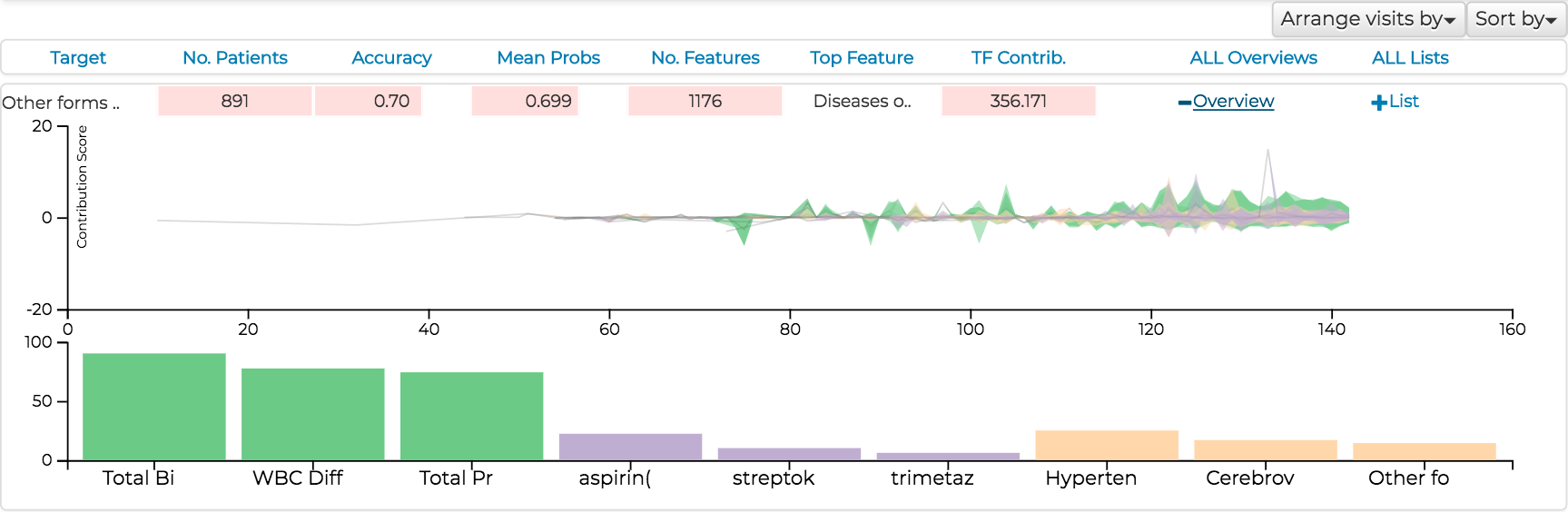}

\vspace{-.07in}
\caption[patient summary]{\emph{\patientsummary} shows a summary of selected patients. Table summarizes description of selected patients. In the middle, an area chart shows aggregated contribution scores of nine medical codes over time. It shows mean and standard deviation as an area. Users can also see the medical codes and their mean contribution scores in bar chart.}
\label{fig:patientsummary}
\vspace{-.2in}
\end{figure}

\patientsummary, in contrast to \overview, shows a temporal summary in contribution progress chart (\textbf{T3}).
There are three charts vertically shown from top to bottom in \patientsummary \bc{as Figure~\ref{fig:patientsummary} shows}.
The first chart is a table that summarizes selected patients: 1) number of patients, 2) accuracy (number of correct prediction / number of patients), 3) mean predicted diagnosis risk, 4) number of medical codes, 5) name of top contributing medical code, and 6) sum of contribution scores. Then, it provides an interaction handle that toggles contribution progress chart and code bar chart.

In contribution progress chart, users can see a temporal overview of nine selected medical contribution scores over sequences or time.
The temporal area chart is constructed in the following way: 1) we align all sequences of medical codes to the final visit; 2) starting from the final visit backward, we compute the mean and standard deviation of contribution scores of the corresponding codes across patients; 3) we visualize the computed means and standard deviation over time as area paths along with the horizontal axis. 
The thickness represents variance, and the vertical spikes show the mean around each visit (with respect to the most recent visit). 
Since patients with longer sequences, such as 120 visits, are rare, it tends to show almost a single line toward the left side.
Figure~\ref{fig:patientsummary} shows that the green codes (diagnosis) show higher variance toward the end of patient records than other types.

Code bar chart shows the top nine contributors of the patients: three per each of three different code types (diagnosis, medication, and disease).
Users can also highlight a code in contribution progress chart by hovering over the corresponding bar.
By clicking on one of the nine codes, users can also sort \patientlist by the contribution scores of it.
The three views provide an overview of selected patients.
After observing peculiar, downward spikes of the contribution scores of aspirin around 10-to-15 visits before the end in Figure~\ref{fig:patientsummary}, users can sort patients by the contribution scores of aspirin in \patientlist (\textbf{T3}).


\vspace{-.05in}
\subsection{\patientlist}
\vspace{-.03in}

\patientlist provides a list of selected patients, where users can explore and compare multiple patients.
In \patientlist, each patient's visit record is represented as rectangular boxes arranged horizontally inspired by prior work in visualizing sequences~\cite{kwon2016visohc, kwon_peekquence_2016}.
Each box decorated with a diverging color scheme of the blue-to-white-to-red (negative-to-0-to-positive) scale represents the sum of contribution scores of all codes in the visit.
At the rightmost end of the visit boxes, a prediction circle icon, which was also used in \overview, shows the strength of the predicted diagnosis risk.
In this view, users can quickly glance the temporal pattern of contribution scores of individual patients and select one patient for a deep-dive analysis.
\bc{Contribution scores show how much each medical code or visit impacts upon prediction certainty. This contribution is the most essential unit of interpretability by showing the relationship between event sequences and predicted outcomes. In the patient list view, we aim to provide when and how much each visit impacted upon high or low prediction certainty for patients. The pattern shown in heatmap is used for users to select interesting patients who have unique patterns.}
Figure~\ref{fig:teaser}~(C) shows a list of patients with a high predicted diagnosis risk.
Patients tend to have visits with high contribution scores towards more recent visits, but exceptions can be seen (\textbf{T5}).
In \patientlist, users can invoke \patientdetail and \patienteditor of a selected patient.



\vspace{-.05in}
\subsection{\patientdetail}
\vspace{-.03in}




\patientdetail shows a focused view of a single patient (\textbf{T4}).
It consists of three different views as Figure~\ref{fig:teaser}~(E) shows.
The first view is a line chart of prediction scores.
The predicted diagnosis risks over time (sequences) are calculated in the following way: 1) starting from the first visit, we predict diagnosis risks by considering only the preceding visits until the corresponding visit; 2) then, we compute \textit{N} predicted diagnosis risks per patient, where \textit{N} is the total number of visits per patient; 3) we also compute the contribution scores of individual medical codes per predicted risk, which will be used in temporal code chart.

Temporal code chart shows contribution scores of all medical codes for each patient.
The view is similarly arranged horizontally per visit as in \patientlist.
\bc{In \patientdetail, we mapped the horizontal space for temporal progression and the vertical space for contribution scores as well as diagnostic progression risks.
In this way, users are able to observe correlation between contribution scores of medical codes and prediction risks.}
Temporal code chart unpacks individual visits into separate medical codes.
The medical codes are represented as colored symbols: green plus (diagnosis), purple diamond (prescription), yellow rectangle (sickness), according to their type.
The colored symbols are placed vertically with respect to their contribution scores.
Code bar chart shows the top nine contributing medical codes.

In \patientdetail, users can understand the progression of predicted diagnosis risks and why such predictions are made (\textbf{T5}).
When users hover over the x-axis, they can see the updated contribution scores of medical codes of preceding visits until the point of time.


\vspace{-.05in}
\subsection{\patienteditor}
\vspace{-.03in}



\patienteditor allows users to conduct what-if analyses (\textbf{T6}).
There are two ways to invoke \patienteditor.
First, users can select a patient in \patientlist, and open a pop-up dialog of \patienteditor (see Figure~\ref{fig:teaser}~(D)).
It provides a dedicated space for editing a selected patient's medical codes.
\patienteditor presents each visit horizontally in a temporal manner and lists each visit's medical codes downward as shown in Figure~\ref{fig:teaser}~(D).
\bc{User can sort medical codes of a visit either by contribution score (default) or by the code type.}
\bc{This layout enables users to easily select medical codes to make changes for interaction.}
Second, users can convert \patientdetail into \patienteditor by simply choosing a context menu option.
By doing so, users can maintain the context while editing the patient visits and medical codes if they were focusing on \patientdetail.
However, users will lose the original version if they directly edit on \patientdetail.
Since there is a tradeoff between the two approaches, we implemented both features and allowed users to choose one at their convenience.

\begin{figure}[t]
\centering


   \includegraphics[width=1\linewidth, frame, trim={0 6cm 0 3.5cm},clip]{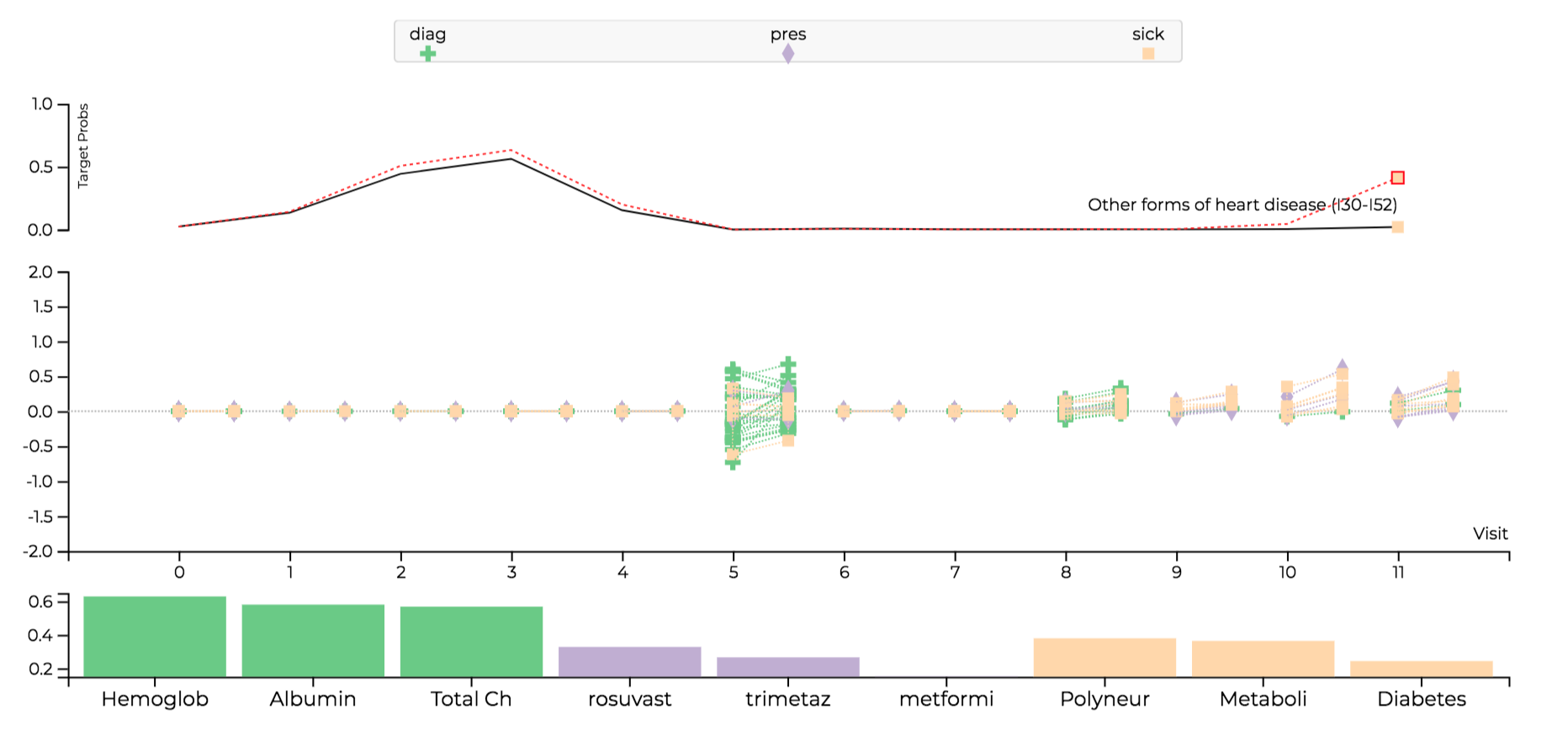}

\caption[patient summary]{The what-if analysis result shows increase in predicted diagnosis risks and contribution scores of related medical codes.}
\label{fig:patienteditorassessment}
\vspace{-.2in}
\end{figure}

As shown in Figure~\ref{fig:teaser}~(D), users can move the visit along the time axis to change the date.
Users can also add new codes in to a visit, and they can remove existing ones.
In some cases, users may feel that they need to steer the model towards their prior knowledge or hypotheses.
In \patienteditor, users can provide feedback to the model (\textbf{T7}) by requesting to increase contribution scores of selected medical codes.
In such activities mentioned above, users can test hypothetical scenarios.
Once users complete the changes, the model returns the newly generated predicted diagnosis risk over time as well as contribution scores overlaid on top of the original records.
For example, users might have felt the need to update contribution scores of selected medical codes and move some visits to different dates (Figure~\ref{fig:teaser}~(D)).
The results are shown in Figure~\ref{fig:patienteditorassessment}.
The predicted risk significantly increased; in particular, predicted risks of final two increased as the red dotted line shows in Figure~\ref{fig:patienteditorassessment}.
The increase was due to increases in contribution scores of medical codes from the four most recent visits, which are shown as the right upward trends in connected code symbols.


\vspace{-.05in}
\section{Experiments}
\label{sec:experiments}
\vspace{-.03in}
This section reports the methods and results of quantitative and qualitative experiments using our models trained on the HIRA-NPS dataset for predicting heart failure and cataract, respectively. 

\vspace{-.05in}
\subsection{Experimental Setup}
\label{sec:experimentalsetup}
\vspace{-.03in}

The primary purpose of RETAIN is to provide interpretatable results
on data-driven medical prediction tasks. To apply our visualization
framework to a case of medical prediction, we set two binary prediction
tasks: predicting a patient's future diagnosis of (a) heart failure
and (b) cataract. Our specific goal is to observe the medical records
of a patients for the first six months to predict whether he/she will
become diagnosed with that condition in the near future. For each
task, we create a case set which consists of patients whose first
diagnosis of the target condition occurred \textit{after}
the end of June. We discard all visits made after June, and remove
patients who has made less than 5 visits. \bc{Both heart failure and cataract datasets showed a similar distribution of patients. For heart failure, the min, max, and average visit lengths were 5, 188, and 20.79. For cataract, these lengths were 5, 148, and 20.05.}

\bc{Each visit of a patient contains a timestamp and the codes related
to the medical treatment, prescriptions, and diagnosed conditions
of a patient. However, the number of codes are too diverse for our
model to properly handle, and thus an additional step for reducing
the total number of codes was taken. For diagnosis codes, we simply
categorized each specific code according to the 268 mid-level
categories according to KCD-9. However, since there were no provided classification schemes for treatment and prescription codes, for each
type of code we selected the n-frequent codes that account for at
least 95\% of the entire data and discarded the rest. We were able
to reduce more than 7,000 treatment codes to 500 and 3,800 prescription
codes to 632 while preserving 94.7\% of the original data. This enables
us to represent all the codes associated with a visit in a 1,400-dimensional
binary vector.}

\bc{For each patient in the
case set, we create a control set consisting of 10 patients who belong
to the same gender and age groups and have a similar number of
visits. We assign target labels of 1 and 0 to the case and control
patients respectively. Thus, each batch contains 11 patients.
We result in 5,730 batches (63,030 patients) for heart failure and 10,692 batches (117,612 patients) for cataract.
We split each dataset into train, validation and test sets with a ratio of 0.65/0.1/0.25.}

Our models are implemented on Pytorch 0.3.1 \cite{paszke2017pytorch}. We trained our model on our training set using Adam \cite{kingma14adam} on learning rates of 0.01, 0.001, 0.0001 and hidden state sizes of 64, 128, 256, and tested them on the validation set to obtain the best performing hyperparameters \bc{and avoid overfitting}. We used an Ubuntu 14.04 server equipped with two Nvidia Titan X GPUs to train our models. \bc{According to our setting, our \retainex model takes 3.19 hours to train using 40,964 patients in our use case for 10 epochs, with 1148 seconds per epoch. For testing, it takes 366 seconds to make predictions of diagnosis risk scores as well as to generate contribution scores of patients' visits and codes. }

\vspace{-.05in}
\subsection{Quantitative analysis}
\vspace{-.03in}
Models are quantitatively evaluated by two metrics; Area under the ROC Curve (AUC) and Average Precision (AP). These measures show robustness to data imbalance in positive/negative labels as they measure how successfully positive cases are ranked above negative cases~\cite{fawcett04roc}.

To further test our model, we implemented two baseline models
for comparison: 1) GRU: We implemented a GRU model using the final hidden state, equivalent to $\boldsymbol{\beta}_{T}$ in our proposed model; 2) 
RETAIN: We implemented the original version of RETAIN.
We also tested the importance of adding time data to \retainex, so we compared it to an equivalent version without time intervals being used (\retainex w/o time).
For the baseline models, we apply the same training and hyperparameter
selection strategy as mentioned above.

\begin{table}
\centering
\caption{Model performances measured for medical predictions tasks}
\label{quant}
\small
\begin{tabular}{|c|c|c|c|c|c|c|}
\hline 
& \multicolumn{3}{c|}{(a) heart failure} & \multicolumn{3}{c|}{(b) cataract}\tabularnewline
\hline 
Models & AUC & AP & time~(s) & AUC & AP & time~(s)\tabularnewline
\hline 
GRU & 0.906 & 0.694 & 997 & 0.953 & 0.834 & 2367\tabularnewline
\hline 
RETAIN & 0.905 & 0.729 & 1114 & 0.959 & 0.835 & 2700\tabularnewline
\hline 
 \retainex w/o time & 0.946 & 0.769 & 1143 & \textbf{0.975} & 0.870 & 2619\tabularnewline
\hline 
\retainex & \textbf{0.954} & \textbf{0.818} & 1148 & \textbf{0.975} & \textbf{0.878} & 2632\tabularnewline
\hline 
\end{tabular}
\label{tab:quant}
\vspace{-.1in}
\end{table}

Table~\ref{tab:quant} shows that \retainex outperforms the baseline models in all cases. The effect of adding an additional embedding matrix can be seen in comparison to the original RETAIN model and \retainex without time. Given the otherwise identical settings, the improvement in performance is due to having two embeddings, one for computing the attention values and another for computing the final prediction output. Furthermore, we show that with the addition of time interval information, the performance of the model increases even more across all settings.
By using the temporal dimension, our model can more accurately learn how to discriminate between important and unimportant visits. \bc{Last of all, the rightmost columns display the average time taken to train the model per epoch, that is to observe every training data once. Table~\ref{tab:quant} shows that RetainEX outperforms baseline models at the expense of only a small increase in training time.}

\vspace{-.05in}
\subsection{Qualitative analysis}
\vspace{-.03in}
We also qualitatively reviewed whether the medical codes (treatment, diagnosis, and prescriptions) with high contribution scores
for predicting heart failure (HF), are supported by general medical knowledge. 
Using the scheme introduced in Section~\ref{sec:model_description}.5,
we generated a score vector $\mathbf{S}$ and an additional 1400-dimensional vector $\mathbf{C}$ for every case patient. $\mathbf{C}$ stores the total counts of each medical code per patient. 
After computing the vectors for every patient,
we sum all $\mathbf{S}$'s and $\mathbf{C}$'s to obtain 1400-dimensional
representations of the contribution scores and counts for the medical
codes of all patients, which we denote as $\mathbf{S}_{total}$
and $\mathbf{C}_{total}$. We normalize $\mathbf{S}_{total}$ in two
different directions: (i) to identify common medical codes prevalent
in most patients, we averaged all dimensions of $\mathbf{S}_{total}$
by $N$ to obtain $\mathbf{S}_{1}$ (Table~\ref{tab:top5featsscores_per_person} in Appendix~\ref{app:qualitative}); (ii) to identify
codes that are strongly associated with the development of heart
failure, we divided each dimension of $\mathbf{S}_{total}$
by its corresponding $\mathbf{C}_{total}$ value to obtain
$\mathbf{S}_{2}$ (see Table~\ref{tab:top5featsscores_per_app} in Appendix~\ref{app:qualitative}).

The top-5 $\mathbf{S}_{1}$ scores in diagnosis support the premise
that hypertensive disease are associated with heart failure,
as well as being a major cause of other diseases and comorbidities
\cite{kannel1972role, levy1996progression}.
Hypertensive disease was the most frequently diagnosed co-morbidity
in patients with heart failure (Table~\ref{tab:top5featsscores_per_person} in Appendix~\ref{app:qualitative}). Likewise, ischaemic heart disease was also a major disease in patients with heart failure \cite{
ozbaran2004autologous}, as reflected by the relatively high $\mathbf{S}_{1}$
scores in the current study. Metabolic disorders, such as hemochromatosis,
for which a relatively high $\mathbf{S}_{1}$ score was observed,
were also shown to be likely to cause heart failure as a complication
\cite{chati1996physical}. It was presumed that cerebrovascular
disease would be diagnosed in a high number of patients with heart
failure as hypertension is a common characteristic of both diseases \cite{shimada1990silent,
caplan1986race}. Bisoprolol, a medicine ingredient for which a relatively
high $\mathbf{S}_{1}$ score was recorded, is frequently prescribed
for heart disease \cite{de2004effects},
while aspirin and atorvastatin are commonly used to prevent it 
\cite{cleland2004warfarin}. 

Carvedilol is a major prescription agent used to treat heart failure \cite{cohn1997safety}.
The results of this study demonstrated that medical codes involved
in the prevention or treatment of heart failure (i.e., prescriptions)
had relatively high scores in $\mathbf{S}_{1}$. Other medical codes with a relatively strong contribution score included obesity, confirming
that it is a major risk factor for heart diseases~\cite{kenchaiah2002obesity}. Disorders of the thyroid
gland are also known to cause heart failure \cite{klein2007thyroid}.
These codes were found to have high scores in $\mathbf{S}_{2}$
(Table~\ref{tab:top5featsscores_per_app} in Appendix~\ref{app:qualitative}). It was assumed that isosorbide mononitrate and amlodipine besylate had relatively high $\mathbf{S}_{2}$ scores because they
are used to treat hypertension, a major causative condition of heart
failure. Heart failure is a clinical syndrome that is characterized
by complicated pathophysiology. 
The results from the study show that our model is capable of identifying factors (i.e., medical codes) that are strongly associated with heart failure.

\vspace{-.05in}
\section{Case Study: Patients with Heart Failure}
\label{sec:casestudy}
\vspace{-.03in}

In this section, we provide a case study, developed and discussed by analyzing a subset of EMR data.
To illustrate the story vividly, we introduce a fictitious character called Jane.
Jane is a data analyst, who is a domain expert in the medical field.
She is very interested in analyzing patients with heart failure (HF) and determining sequences of medical codes that are related to the onset of the disease.

Jane decided to conduct a predictive analysis using \toolname with \retainex trained by 40,964 patients (1:10 ratio between case and control) for the heart failure case study (see Section~\ref{sec:experiments} for details). 
She pulled 3,724 patients diagnosed with heart failure in the latter half of a calendar year.
She then launched \toolname to see an overview of the patients in terms of contribution scores of 1,400 medical codes.

The initial overview showed a very interesting grouping in the upper right corner of \overview (see the highlighted area in Figure~\ref{fig:teaser}~(A)).
Jane filtered patients by drawing a polygon area of interest over the region using the lasso tool.
The initial selection provided 564 patients (F = 297) with very high prediction scores on average (.97), which indicates that the patients are explained well with \retainex.
She loaded the selection into \patientsummary.
It showed the top three contributing diseases (comorbidities) as ischaemic heart disease, hypertensive disease, and cerebrovascular disease, all of which are known to be highly related to heart failures.
In particular, the existence of hypertensive disease indicates its relevance to the S1 General HF group in Yan et al.~\cite{yan2016learning}.
The top three medications are bisoprolol hemifumarate, aspirin, and trimetazidine.
Bisoprolo is related to reducing hypertension, and trimetazidine is related to ischaemic heart disease.
It was interesting to see aspirin among the top contributors as it is known to reduce the HF risk with potential side effects like kidney failure for long-term use.

Jane quickly broke down the group into a more granular level.
The data points were subdivided into three subgroups, each of which tends to be cohesive within its group but separated from others.
The first subgroup (N$=$201) showed the similar representation of what we saw with high hypertension and Bisoprolo (S1).
The second group showed an interesting diagnosis called ``syndrome of four (or Sasang) constitutional medicine'' as one of the high contributing medical codes.
The Sasang typology is a traditional Korean personalized medicine, which aims to cluster patients into four groups based on patient's phenotypic characteristics~\cite{chae2017personalized}.
It was interesting to observe that a fair number of patients (N$=$230) showed the influence of this unique medical code.
Recently, there have been studies investigating the relationship between the Sasang types and prediction of cardiovascular disease~\cite{cho2013relationship}.
She thought it will be interesting to test such hypotheses later.

The third group showed another interesting cohort with relatively higher age (74.7 years old in average) than the other two groups (66.7 years old).
In \patientsummary, Jane saw that the group is associated with hypertension and diseases of oesophagus.
It has been reported that there might be relationship between heart disease (e.g., ischaemic heart disease) and diseases of oesophagus (e.g., Barrett's esophagus)~\cite{tsibouris2014ischemic}.
The group also showed high contribution scores of bilirubin, suspected as a predictive marker of pulmonary arterial hypertension~\cite{takeda2010bilirubin}.
She conjectured that this group shows many severe diseases (mostly related to high blood pressure) with high prediction scores.

Jane decided to drill down into details in \patientlist.
She sorted the list by the number of visits then hovered over cerebrovascular disease (the top contributor of this group), and selected a patient with a very high volume of visits (N$=$150) over the period of six months.
She observed cerebrovascular diseases recorded for almost every visit.
By pulling the patient's detail in \patientdetail, she found that the patient is taking a variety of preventive medicines as top three contributors: glimepiride (anti-diabetic drug), pravastatin (prevent high cholesterol), and hydrochlorothiazide (prevent high blood pressure). 
By arranging the x-axis by dates, she also realized that the patient was prescribed the medicine periodically (once in every two weeks). 
The patient was also diagnosed with metabolic disorders nearly every visit.
In summary, Jane could confirm many known stories about heart failures, where it is closely related to metabolic disorders, hypertension, and growing age.

Jane switched her gear to evaluate the performance of the predictive model.
Since she in general believed that the model describes the heart failure prediction very well with associated comorbidities and medication as high contribution scores, she was curious of cases where the model failed.
Could it be due to the data quality?
She sorted the patients by prediction scores, and found three patients who were not predicted as HF (prediction score $<$ .5).
She selected a patient with the lowest score (.076).
Interestingly, this patient did show the prevalence of aforementioned medical codes, such as hypertention, bilirubin, and aspirin, towards the end of June.
However, there was a very unique aspect of this patient.
There were major injuries recorded in May 20, namely head, leg, body injuries
leading to medication prescriptions related to pain (e.g., tramadol) on next two visits.
Also, the patient was diagnosed with arthrosis twice.
Jane conjectured the mixture of major injuries with HF related diseases might be the issue.
She promptly conducted a what-if analysis using \patienteditor.
She removed injuries and related medications, and tightened dates between events towards the end of June.
She selected hypertension, bilirubin, aspirin, and ischaemic heart disease, then chose the ``increase the contribution score'' option to retrain the model.
The retrained model with new input increased the prediction score from .076 to .600 with hypertension as the highest contributor.
She hypothesized that it will be difficult to perfectly predict HF when a patient is associated with parallel activities.
She once again realized the danger of purely automatic solutions and the importance of collaboration between human and machine via visual analytics.



    

\vspace{-.025in}
\section{Discussion}
\label{sec:discussions}
\vspace{-.03in}
In this section, we provide an in-depth discussion of our study by sharing lessons for designing visual analytics applications using RNN-based models on a diagnostic risk prediction task. 

\vspace{-.025in}
\subsection{Interpretability and Model Performance}
\vspace{-.03in}


Adding interpretability to a model while preserving its performance is a challenging task. The strength of RNNs comes from its intertwined structure that freely learns high-dimensional representations in the latent space provided by its hidden states. In this sense, our approach of improving interpretability using linear combinations can be seen as forcing the model to learn these representations at the expense of computational freedom.
Thus, understanding the tradeoff between interpretabiltiy and the performance of RNN models is crucial in designing a visual analytics tool. Target user tasks can be a guidance to solving this deadlock. Our tasks in Section~\ref{sec:tasks} show an example.

One golden rule of data visualization is to maintain simplicity, which we discovered applies to the case of medical data as well. One expert expressed his thoughts of an ideal visualization tool for EMR data as a `conversation partner' and that the first step to fulfilling this role is to have visualization results as simple as possible. He also revealed that it is important for the model and results to be intepretable and interactive, but also easily explainable by design.
He also pointed out clinicians may not benefit much from the complex information provided by a machine. The contribution scores of each code and the predicted outcome presented by the model are, in a doctor's eyes, an additional piece of information that has to be looked at and verified. This might actually put extra burden to the doctor and hinder the process of decision making, instead of assisting it. This opinion was supported by another expert stating that visualization methods were partly unintuitive and difficult to interpret at first sight. This was contrary to our belief that presenting more information would lead to more precise diagnoses.

Both experts agreed that a more welcoming situation would be to develop a machine that simplifies the already complicated EMR data and pinpoints a number of points of attention, similar to our visit-level attention view. It is not a meticulous analysis tool that domain experts need, but an agent that can suggest interesting discussion points by looking at the data from a different perspective, just as if it were a fellow expert. Of course, the machine should be able to sufficiently prove why it made such a prediction or emphasized on a particular visit of a patient, so interpretability still remains a prerequisite. However, that is only when the user asks the model to prove its prediction, and in general the level of visualization should remain as simple as possible.

This feedback led us to different variations of our model for different purposes. The current complexity of our tool can be used to aid researchers who would like to freely explore the available data and conduct various what-if case analyses as seen in Section~\ref{sec:casestudy}. Meanwhile, a more simplified version highlighting only significant and anomalous events can be adopted as an assistant tool for clinicians. The presented events may provide new insights that might have otherwise been overlooked. Thus, designers and providers need to correctly identify target users' needs and maximize the desired effects out of the given settings.

\vspace{-.025in}
\subsection{Towards Interactivity}
\vspace{-.03in}
Another important objective of our work was to apply user interaction schemes for various what-if analyses. To allow for a greater depth of user interaction, we added the following functions to the original model: (1) we used the interval information between visits as an additional input feature to our model, and (2) we introduced a retraining method using an additional embedding matrix to increase or decrease the contributions of individual codes according to the domain knowledge of the user. We also showed that not only do these additional functions ensure our proposed interaction features, but they have an auxiliary effect of improving the model's quantitative performance as well.

While making use of temporal data is also an important concept that we present, here we focus more on the retraining module. This strategy was effective in correcting the contribution scores that were assigned to certain visits or medical codes. We selected a case patient and a control patient, who were misclassified with the model trained initially (Type-II and Type-I errors). With the help of medical experts, we found which codes were over- and under-represented and updated their contribution scores accordingly. Not only were we able to fix the prediction scores of the selected samples, but we also noticed that the mean diagnostic risk prediction scores of a test dataset increased from 0.812 to 0.814. That is, our retraining scheme conducted at one or two samples ended up improving the overall performance of the model without affecting the model's integrity in computing attention scores for other samples.

The fact that retraining the contributions of patients leads to improved performance shows an example of how users can teach the machine based on their domain knowledge. Such interactions can resolve inherent problems of machine learning-based models, where the performance does not improve without additional training data.
In particular, \toolname retrained the model using users' updates made to contribution scores of medical codes.
In other words, the feature-level interpretation can be used as an interaction handle.
In this manner, users do not have to directly update the model parameters, which can be challenging for domain experts.
This study illustrates one way to reflect user's intent, namely using direct manipulation and menu selection on feature-level representation of data points.

    

\vspace{-.025in}
\subsection{Issues in Visualization and AI for Health}
\vspace{-.03in}



A major concern is the risk of the machine making false predictions. No matter how accurate a diagnosis prediction model may be, there is always the possibility that it will produce Type-II errors and fail to capture a serious condition of a patient in life-or-death problems. Thus, solely relying on the information provided by a machine becomes risky because doctors have to take full responsibility of a patient's outcome. In addition, the performance metrics need to be more convincing. Though AUC and AP are proven to be effective metrics for measuring the performances in imbalanced datasets, a high score does not necessarily mean that a model makes a clear distinction between safe and suspected patients. While an ideal situation would be to have a threshold around 0.5 out of 1 to discriminate between positive and negative cases, we discovered that the threshold that maximizes the F-1 score of the predicted results was relatively low, near 0.2. This reflects a common problem in applying machine learning to medical prediction tasks, where a high score cannot guarantee to prevent a serious mistake.

Visualizations can be carefully used to communicate the performance of the model in a transparent way.
Domain experts often hesitate to accept what they see as facts since they are unable to tell various uncertainties propagated through the pipeline~\cite{sacha_role_2016}.
There might be artificial patterns created due to inconsistent EMR recording practice, which could then be amplified by incorrectly trained models. Even for examples that were proven to be accurate, visualizations should indicate its inherent uncertainties involved. Thus, future researchers can investigate the design of uncertainty visualizations, when applying deep learning-based models for complex medical data.

Additionally, we learned that different tasks of the medical domain have to be modeled differently to produce satisfactory results. Another medical task that we did not include in this work is that of predicting the main sickness of a patient's next visit. Though the same settings were used, we discovered that even using the same number of input and output classes as in our proposed setting, RETAIN failed to outperform even the simplest baseline that returns the most frequent diagnosed sickness of each patient. While such problems can be left for future work, we would like to emphasize that in order for a machine learning-based model to prevail, a substantial amount of time and effort are required to tailor the problem setting and preprocess given data.


\vspace{-0.015in}
\subsection{Limitations and Future Work}
\label{sec:scalability}
\vspace{-0.03in}
\bc{
Our tool has several limitations in terms of scalability due to its computationally expensive deep learning-based model and limited screen real estates. Since the scatter plot can suffer from overplotting given too many patients, our recommendation is not to go beyond 10K patients visualized in the scatter plot. 
The Lasso-selection in the scatter plot can mislead users as described in Section~\ref{sec:tool_description}. 
Our line charts may not work when we visualize more than decades of records with more than thousands of visits. In such a case, we recommend to use a horizontal scroll for users to navigate through users' medical history. Our main interactions to view updated results based on changes the user made to input values, can be improved. We may compute and visualize all possible input value combinations that can lead to the improvement of outcomes. Though our case study demonstrates the usefulness of \toolname, it has not been empirically tested whether and how various features of \toolname can help users solve problems. Our future work aims to conduct a long-term user study, where health care professionals will provide feedback to improve the overall design and usability issues. 
}

\vspace{-0.15in}
\section{Conclusion}
\label{sec:conclusion}

In this study, we developed a visual analytics system called \toolname by incorporating \retainex into electronic medical datasets.
Our iterative design process led us to improve interpretability as well as interactivity while maintaining its performance level against RETAIN.
Our study shows that the design of \toolname helps users explore real-world EMRs, gain insights, and generate new hypotheses.
We aim to extend our approach to more diverse medical records, including various measures from medical tests, sensor data from medical equipment and personal devices.
We believe that the lessons learned from this study can better guide future researchers to build interpretable and interactive visual analytics systems for recurrent neural network models.

\acknowledgments{
We thank Wonkyu Kim, who participated in discussion to improve the design of \toolname, and our colleagues from IBM Research, Korea University, Georgia Institute of Technology, and other institutions, who provided constructive feedback. This research was partly supported by Korea Electric Power Corporation (Grant Number: R18XA05).
}

\newpage

\bibliographystyle{abbrv-doi}

\bibliography{Retain}

\begin{thebibliography}{10}

\bibitem{araujo17cancer}
T.~Araújo, G.~Aresta, E.~Castro, J.~Rouco, P.~Aguiar, C.~Eloy, A.~Polónia,
  and A.~Campilho.
\newblock Classification of breast cancer histology images using convolutional
  neural networks.
\newblock {\em PLOS ONE}, 12(6):1--14, 06 2017.

\bibitem{aupetit_multidimensional_2014}
M.~Aupetit, N.~Heulot, and J.-D. Fekete.
\newblock A multidimensional brush for scatterplot data analytics.
\newblock In {\em Visual Analytics Science and Technology (VAST)}, pp. 221 --
  222. IEEE, Oct. 2014.

\bibitem{bahdanau14attention}
D.~Bahdanau, K.~Cho, and Y.~Bengio.
\newblock Neural machine translation by jointly learning to align and
  translate.
\newblock {\em International Conference on Learning Representations}, 2015.

\bibitem{baytas17timelstm}
I.~M. Baytas, C.~Xiao, X.~Zhang, F.~Wang, A.~K. Jain, and J.~Zhou.
\newblock Patient subtyping via time-aware lstm networks.
\newblock In {\em Proceedings of the ACM SIGKDD International Conference on
  Knowledge Discovery and Data Mining}, pp. 65--74, 2017.

\bibitem{caplan1986race}
L.~Caplan, P.~Gorelick, and D.~Hier.
\newblock Race, sex and occlusive cerebrovascular disease: a review.
\newblock {\em Stroke}, 17(4):648--655, 1986.

\bibitem{cavallo2017exploring}
M.~Cavallo and {\c{C}}.~Demiralp.
\newblock Exploring dimensionality reductions with forward and backward
  projections.
\newblock {\em arXiv preprint arXiv:1707.04281}, 2017.

\bibitem{chae2017personalized}
H.~Chae, J.~Lee, E.~S. Jeon, and J.~K. Kim.
\newblock Personalized acupuncture treatment with sasang typology.
\newblock {\em Integrative Medicine Research}, 6(4):329--336, 2017.

\bibitem{chati1996physical}
Z.~Chati, F.~Zannad, C.~Jeandel, B.~Lherbier, J.-M. Escanye, J.~Robert, and
  E.~Aliot.
\newblock Physical deconditioning may be a mechanism for the skeletal muscle
  energy phosphate metabolism abnormalities in chronic heart failure.
\newblock {\em American Heart Journal}, 131(3):560--566, 1996.

\bibitem{che15phenotyping}
Z.~Che, D.~Kale, W.~Li, M.~T. Bahadori, and Y.~Liu.
\newblock Deep computational phenotyping.
\newblock In {\em Proceedings of the ACM SIGKDD International Conference on
  Knowledge Discovery and Data Mining}, pp. 507--516, 2015.

\bibitem{cho2014gru}
K.~Cho, B.~van Merrienboer, {\c{C}}.~G{\"{u}}l{\c{c}}ehre, D.~Bahdanau,
  F.~Bougares, H.~Schwenk, and Y.~Bengio.
\newblock Learning phrase representations using {RNN} encoder-decoder for
  statistical machine translation.
\newblock In {\em Proceedings of the 2014 Conference on Empirical Methods in
  Natural Language Processing}, pp. 1724--1734, 2014.

\bibitem{cho2013relationship}
N.~H. Cho, J.~Y. Kim, S.~S. Kim, and C.~Shin.
\newblock The relationship of metabolic syndrome and constitutional medicine
  for the prediction of cardiovascular disease.
\newblock {\em Diabetes \& Metabolic Syndrome: Clinical Research \& Reviews},
  7(4):226--232, 2013.

\bibitem{choi16doctorai}
E.~Choi, M.~T. Bahadori, A.~Schuetz, W.~F. Stewart, and J.~Sun.
\newblock Doctor ai: Predicting clinical events via recurrent neural networks.
\newblock In {\em Proceedings of the 1st Machine Learning for Healthcare
  Conference}, vol.~56, pp. 301--318, 2016.

\bibitem{choi16med2vec}
E.~Choi, M.~T. Bahadori, E.~Searles, C.~Coffey, M.~Thompson, J.~Bost,
  J.~Tejedor{-}Sojo, and J.~Sun.
\newblock Multi-layer representation learning for medical concepts.
\newblock In {\em Proceedings of the 22nd {ACM} {SIGKDD} International
  Conference on Knowledge Discovery and Data Mining}, pp. 1495--1504, 2016.

\bibitem{choi17gram}
E.~Choi, M.~T. Bahadori, L.~Song, W.~F. Stewart, and J.~Sun.
\newblock {GRAM:} graph-based attention model for healthcare representation
  learning.
\newblock In {\em Proceedings of the 23rd {ACM} {SIGKDD} International
  Conference on Knowledge Discovery and Data Mining}, pp. 787--795, 2017.

\bibitem{choi2016retain}
E.~Choi, M.~T. Bahadori, J.~Sun, J.~Kulas, A.~Schuetz, and W.~Stewart.
\newblock Retain: An interpretable predictive model for healthcare using
  reverse time attention mechanism.
\newblock In {\em Advances in Neural Information Processing Systems 29}, pp.
  3504--3512. Curran Associates, Inc., 2016.

\bibitem{choi17heart}
E.~Choi, A.~Schuetz, W.~F. Stewart, and J.~Sun.
\newblock Using recurrent neural network models for early detection of heart
  failure onset.
\newblock {\em {Journal of the American Medical Informatics Association}},
  24(2):361--370, 2017.

\bibitem{choo13utopian}
J.~Choo, C.~Lee, C.~K. Reddy, and H.~Park.
\newblock Utopian: User-driven topic modeling based on interactive nonnegative
  matrix factorization.
\newblock {\em IEEE Transactions on Visualization and Computer Graphics},
  19(12):1992--2001, Dec 2013.

\bibitem{choo15cluster}
J.~Choo, C.~Lee, C.~K. Reddy, and H.~Park.
\newblock Weakly supervised nonnegative matrix factorization for user-driven
  clustering.
\newblock {\em Data Mining and Knowledge Discovery}, 29(6):1598--1621, Nov
  2015.

\bibitem{chung2016re}
S.~Chung, C.~Park, S.~Suh, K.~Kang, J.~Choo, and B.~C. Kwon.
\newblock {Re-VACNN}: Steering convolutional neural network via real-time
  visual analytics.
\newblock In {\em Future of Interactive Learning Machines Workshop at the 30th
  Annual Conference on Neural Information Processing Systems}, 2016.

\bibitem{cleland2004warfarin}
J.~Cleland, I.~Findlay, S.~Jafri, G.~Sutton, R.~Falk, C.~Bulpitt, C.~Prentice,
  I.~Ford, A.~Trainer, and P.~Poole-Wilson.
\newblock The warfarin/aspirin study in heart failure (wash): a randomized
  trial comparing antithrombotic strategies for patients with heart failure.
\newblock {\em American Heart Journal}, 148(1):157--164, 2004.

\bibitem{cohn1997safety}
J.~N. Cohn, M.~B. Fowler, M.~R. Bristow, W.~S. Colucci, E.~M. Gilbert,
  V.~Kinhal, S.~K. Krueger, T.~Lejemtel, K.~A. Narahara, M.~Packer, et~al.
\newblock Safety and efficacy of carvedilol in severe heart failure.
\newblock {\em Journal of Cardiac Failure}, 3(3):173--179, 1997.

\bibitem{de2004effects}
P.~De~Groote, P.~Delour, N.~Lamblin, J.~Dagorn, C.~Verkindere, E.~Tison,
  A.~Millaire, and C.~Bauters.
\newblock Effects of bisoprolol in patients with stable congestive heart
  failure.
\newblock {\em Annales de Cardiologie et d'Angeiologie}, 53(4):167--170, 2004.

\bibitem{ding18visnmt}
Y.~Ding, Y.~Liu, H.~Luan, and M.~Sun.
\newblock Visualizing and understanding neural machine translation.
\newblock In {\em Proceedings of the 55th Annual Meeting of the Association for
  Computational Linguistics}, 2017.

\bibitem{ehrenberg16data}
H.~R. Ehrenberg, J.~Shin, A.~J. Ratner, J.~A. Fries, and C.~R{\'e}.
\newblock Data programming with ddlite: Putting humans in a different part of
  the loop.
\newblock In {\em Proceedings of the Workshop on Human-In-the-Loop Data
  Analytics}, pp. 13:1--13:6. ACM, 2016.

\bibitem{elassady18topic}
M.~El-Assady, R.~Sevastjanova, F.~Sperrle, D.~Keim, and C.~Collins.
\newblock Progressive learning of topic modeling parameters: A visual analytics
  framework.
\newblock {\em IEEE Transactions on Visualization and Computer Graphics},
  24(1):382--391, Jan 2018.

\bibitem{fawcett04roc}
T.~Fawcett.
\newblock Roc graphs: Notes and practical considerations for researchers.
\newblock {\em Machine learning}, 31(1):1--38, 2004.

\bibitem{friedman2001greedy}
J.~H. Friedman.
\newblock Greedy function approximation: a gradient boosting machine.
\newblock {\em Annals of statistics}, pp. 1189--1232, 2001.

\bibitem{han17cancer}
Z.~Han, B.~Wei, Y.~Zheng, Y.~Yin, K.~Li, and S.~Li.
\newblock Breast cancer multi-classification from histopathological images with
  structured deep learning model.
\newblock {\em Scientific Reports}, 7(1):4172, 2017.

\bibitem{havaei17brain}
M.~Havaei, A.~Davy, D.~Warde-Farley, A.~Biard, A.~Courville, Y.~Bengio, C.~Pal,
  P.-M. Jodoin, and H.~Larochelle.
\newblock Brain tumor segmentation with deep neural networks.
\newblock {\em Medical Image Analysis}, 35:18 -- 31, 2017.

\bibitem{heimerl12classifier}
F.~Heimerl, S.~Koch, H.~Bosch, and T.~Ertl.
\newblock Visual classifier training for text document retrieval.
\newblock {\em IEEE Transactions on Visualization and Computer Graphics},
  18(12):2839--2848, Dec 2012.

\bibitem{heulot_proxilens_2013}
N.~Heulot, M.~Aupetit, and J.-D. Fekete.
\newblock {ProxiLens}: {Interactive} {Exploration} of {High}-{Dimensional}
  {Data} using {Projections}.
\newblock In {\em VAMP: EuroVis Workshop on Visual Analytics using
  Multidimensional Projections}. The Eurographics Association, June 2013.

\bibitem{hochreiter1997lstm}
S.~Hochreiter and J.~Schmidhuber.
\newblock Long short-term memory.
\newblock {\em Neural Computation}, 9(8):1735--1780, Nov 1997.

\bibitem{hohman17deep}
F.~Hohman, N.~O. Hodas, and D.~H. Chau.
\newblock Shapeshop: Towards understanding deep learning representations via
  interactive experimentation.
\newblock In {\em Proceedings of the 2017 {CHI} Conference on Human Factors in
  Computing Systems}, pp. 1694--1699, 2017.

\bibitem{jin18heart}
B.~Jin, C.~Che, Z.~Liu, S.~Zhang, X.~Yin, and X.~Wei.
\newblock Predicting the risk of heart failure with ehr sequential data
  modeling.
\newblock {\em IEEE Access}, 6:9256--9261, 2018.

\bibitem{kahng18activis}
M.~Kahng, P.~Y. Andrews, A.~Kalro, and D.~H.~P. Chau.
\newblock Activis: Visual exploration of industry-scale deep neural network
  models.
\newblock {\em IEEE Transactions on Visualization and Computer Graphics},
  24(1):88--97, 2018.

\bibitem{kale15phenotyping}
D.~C. Kale, Z.~Che, M.~T. Bahadori, W.~Li, Y.~Liu, and R.~C. Wetzel.
\newblock Causal phenotype discovery via deep networks.
\newblock In {\em American Medical Informatics Association Annual Symposium},
  2015.

\bibitem{kamnitsas17brain}
K.~Kamnitsas, C.~Ledig, V.~F.~J. Newcombe, J.~P. Simpson, A.~D. Kane, D.~K.
  Menon, D.~Rueckert, and B.~Glocker.
\newblock Efficient multi-scale 3d cnn with fully connected crf for accurate
  brain lesion segmentation.
\newblock {\em Medical Image Analysis}, 36:61--78, 2017.

\bibitem{kannel1972role}
W.~B. Kannel, W.~P. Castelli, P.~M. McNamara, P.~A. McKee, and M.~Feinleib.
\newblock Role of blood pressure in the development of congestive heart
  failure: the framingham study.
\newblock {\em New England Journal of Medicine}, 287(16):781--787, 1972.

\bibitem{karpathy15rnn}
A.~Karpathy, J.~Johnson, and F.~Li.
\newblock Visualizing and understanding recurrent networks.
\newblock {\em CoRR}, abs/1506.02078, 2015.

\bibitem{kenchaiah2002obesity}
S.~Kenchaiah, J.~C. Evans, D.~Levy, P.~W. Wilson, E.~J. Benjamin, M.~G. Larson,
  W.~B. Kannel, and R.~S. Vasan.
\newblock Obesity and the risk of heart failure.
\newblock {\em New England Journal of Medicine}, 347(5):305--313, 2002.

\bibitem{kim2014hira}
L.~Kim, J.-A. Kim, and S.~Kim.
\newblock A guide for the utilization of health insurance review and assessment
  service national patient samples.
\newblock {\em Epidemiology and Health}, 36:e2014008, 2014.

\bibitem{kim2013hira}
L.~Kim, J.~Sakong, Y.~Kim, S.~Kim, S.~Kim, B.~Tchoe, H.~Jeong, and T.~Lee.
\newblock Developing the inpatient sample for the national health insurance
  claims data.
\newblock {\em Health Policy and Management}, 23(2):152--161, Jun 2013.

\bibitem{kingma14adam}
D.~P. Kingma and J.~Ba.
\newblock Adam: A method for stochastic optimization.
\newblock In {\em International Conference on Learning Representations}, 2015.

\bibitem{klein2007thyroid}
I.~Klein and S.~Danzi.
\newblock Thyroid disease and the heart.
\newblock {\em Circulation}, 116(15):1725--1735, 2007.

\bibitem{kwon18cluster}
B.~C. Kwon, B.~Eysenbach, J.~Verma, K.~Ng, C.~D. Filippi, W.~F. Stewart, and
  A.~Perer.
\newblock Clustervision: Visual supervision of unsupervised clustering.
\newblock {\em IEEE Transactions on Visualization and Computer Graphics},
  24(1):142--151, Jan 2018.

\bibitem{kwon17axisketcher}
B.~C. Kwon, H.~Kim, E.~Wall, J.~Choo, H.~Park, and A.~Endert.
\newblock Axisketcher: Interactive nonlinear axis mapping of visualizations
  through user drawings.
\newblock {\em IEEE Transactions on Visualization and Computer Graphics},
  23(1):221--230, Jan 2017.

\bibitem{kwon2016visohc}
B.~C. Kwon, S.-H. Kim, S.~Lee, J.~Choo, J.~Huh, and J.~S. Yi.
\newblock Visohc: Designing visual analytics for online health communities.
\newblock {\em IEEE Transactions on Visualization and Computer Graphics},
  22(1):71--80, 2016.

\bibitem{kwon_peekquence_2016}
B.~C. Kwon, J.~Verma, and A.~Perer.
\newblock {Peekquence}: Visual analytics for event sequence data.
\newblock {\em ACM SIGKDD Workshop on Interactive Data Exploration and
  Analytics}, 2016.

\bibitem{lee12cluster}
H.~Lee, J.~Kihm, J.~Choo, J.~T. Stasko, and H.~Park.
\newblock ivisclustering: An interactive visual document clustering via topic
  modeling.
\newblock {\em Computer Graphics Forum}, 31(3):1155--1164, 2012.

\bibitem{lee17topic}
T.~Y. Lee, A.~Smith, K.~Seppi, N.~Elmqvist, J.~Boyd-Graber, and L.~Findlater.
\newblock The human touch: How non-expert users perceive, interpret, and fix
  topic models.
\newblock {\em International Journal of Human-Computer Studies}, 105:28 -- 42,
  2017.

\bibitem{lespinats_checkviz_2013}
S.~Lespinats and M.~Aupetit.
\newblock {CheckViz}: {Sanity} {Check} and {Topological} {Clues} for {Linear}
  and {Non}-{Linear} {Mappings}.
\newblock {\em Computer Graphics Forum}, 30(1):113--125, 2010.

\bibitem{levy1996progression}
D.~Levy, M.~G. Larson, R.~S. Vasan, W.~B. Kannel, and K.~K. Ho.
\newblock The progression from hypertension to congestive heart failure.
\newblock {\em Journal of the American Medical Association},
  275(20):1557--1562, 1996.

\bibitem{lin18rclens}
H.~Lin, S.~Gao, D.~Gotz, F.~Du, J.~He, and N.~Cao.
\newblock Rclens: Interactive rare category exploration and identification.
\newblock {\em IEEE Transactions on Visualization and Computer Graphics},
  PP(99):1--1, 2018.

\bibitem{lipton2016mythos}
Z.~C. Lipton.
\newblock The mythos of model interpretability.
\newblock {\em arXiv preprint arXiv:1606.03490}, 2016.

\bibitem{lipton15diag}
Z.~C. Lipton, D.~C. Kale, C.~Elkan, and R.~C. Wetzel.
\newblock Learning to diagnose with lstm recurrent neural networks.
\newblock {\em International Conference on Learning Representations}, 2015.

\bibitem{lipton15phenotyping}
Z.~C. Lipton, D.~C. Kale, and R.~C. Wetzel.
\newblock Phenotyping of clinical time series with {LSTM} recurrent neural
  networks.
\newblock In {\em Workshop on Machine Learning in Healthcare at the 29th Annual
  Conference on Neural Information Processing Systems}, 2015.

\bibitem{liu17cnn}
M.~Liu, J.~Shi, Z.~Li, C.~Li, J.~Zhu, and S.~Liu.
\newblock Towards better analysis of deep convolutional neural networks.
\newblock {\em IEEE Transactions on Visualization and Computer Graphics},
  23(1):91--100, 2017.

\bibitem{luong15attention}
T.~Luong, H.~Pham, and C.~D. Manning.
\newblock Effective approaches to attention-based neural machine translation.
\newblock In {\em Proceedings of the 2015 Conference on Empirical Methods in
  Natural Language Processing}, pp. 1412--1421, 2015.

\bibitem{ma2017dipole}
F.~Ma, R.~Chitta, J.~Zhou, Q.~You, T.~Sun, and J.~Gao.
\newblock Dipole: Diagnosis prediction in healthcare via attention-based
  bidirectional recurrent neural networks.
\newblock In {\em Proceedings of the ACM SIGKDD International Conference on
  Knowledge Discovery and Data Mining}, pp. 1903--1911, 2017.

\bibitem{maaten2008visualizing}
L.~v.~d. Maaten and G.~Hinton.
\newblock Visualizing data using t-sne.
\newblock {\em JMLR}, 9(Nov), 2008.

\bibitem{ming17rnnvis}
Y.~Ming, S.~Cao, R.~Zhang, Z.~Li, Y.~Chen, Y.~Song, and H.~Qu.
\newblock Understanding hidden memories of recurrent neural networks.
\newblock In {\em IEEE Conference on Visual Analytics Science and Technology},
  2017.

\bibitem{ozbaran2004autologous}
M.~Ozbaran, S.~B. Omay, S.~Nalbantgil, H.~Kultursay, K.~Kumanlioglu, D.~Nart,
  and E.~Pektok.
\newblock Autologous peripheral stem cell transplantation in patients with
  congestive heart failure due to ischemic heart disease.
\newblock {\em European Journal of Cardio-Thoracic Surgery}, 25(3):342--350,
  2004.

\bibitem{paszke2017pytorch}
A.~Paszke, S.~Gross, S.~Chintala, G.~Chanan, E.~Yang, Z.~DeVito, Z.~Lin,
  A.~Desmaison, L.~Antiga, and A.~Lerer.
\newblock Automatic differentiation in pytorch.
\newblock In {\em The Future of Gradient-based Machine Learning Software and
  Techniques Workshop at the 31st Annual Conference on Neural Information
  Processing Systems}, 2017.

\bibitem{pezzotti18deepeyes}
N.~Pezzotti, T.~Höllt, J.~V. Gemert, B.~P.~F. Lelieveldt, E.~Eisemann, and
  A.~Vilanova.
\newblock Deepeyes: Progressive visual analytics for designing deep neural
  networks.
\newblock {\em IEEE Transactions on Visualization and Computer Graphics},
  24(1):98--108, Jan 2018.

\bibitem{prakash17diag}
A.~Prakash, S.~Zhao, S.~A. Hasan, V.~V. Datla, K.~Lee, A.~Qadir, J.~Liu, and
  O.~Farri.
\newblock Condensed memory networks for clinical diagnostic inferencing.
\newblock In {\em Proceedings of the Thirty-First {AAAI} Conference on
  Artificial Intelligence}, pp. 3274--3280, 2017.

\bibitem{rauber17vis}
P.~E. Rauber, S.~G. Fadel, A.~X. Falcão, and A.~C. Telea.
\newblock Visualizing the hidden activity of artificial neural networks.
\newblock {\em IEEE Transactions on Visualization and Computer Graphics},
  23(1):101--110, Jan 2017.

\bibitem{ribeiro2016should}
M.~T. Ribeiro, S.~Singh, and C.~Guestrin.
\newblock Why should i trust you?: Explaining the predictions of any
  classifier.
\newblock In {\em Proceedings of the 22nd ACM SIGKDD International Conference
  on Knowledge Discovery and Data Mining}, pp. 1135--1144. ACM, 2016.

\bibitem{sacha17vis}
D.~Sacha, M.~Sedlmair, L.~Zhang, J.~A. Lee, J.~Peltonen, D.~Weiskopf, S.~C.
  North, and D.~A. Keim.
\newblock What you see is what you can change: Human-centered machine learning
  by interactive visualization.
\newblock {\em Neurocomputing}, 268:164 -- 175, 2017.

\bibitem{sacha_role_2016}
D.~Sacha, H.~Senaratne, B.~C. Kwon, G.~Ellis, and D.~A. Keim.
\newblock The {Role} {Of} {Uncertainty}, {Awareness}, {And} {Trust} {In}
  {Visual} {Analytics}.
\newblock {\em Visualization and Computer Graphics, IEEE Transactions on},
  22(1):240--249, 2016.

\bibitem{sedlmair2012design}
M.~Sedlmair, M.~Meyer, and T.~Munzner.
\newblock Design study methodology: Reflections from the trenches and the
  stacks.
\newblock {\em IEEE Transactions on Visualization and Computer Graphics},
  18(12):2431--2440, 2012.

\bibitem{shapley1953value}
L.~S. Shapley.
\newblock A value for n-person games.
\newblock {\em Contributions to the Theory of Games}, 2(28):307--317, 1953.

\bibitem{shimada1990silent}
K.~Shimada, A.~Kawamoto, K.~Matsubayashi, and T.~Ozawa.
\newblock Silent cerebrovascular disease in the elderly. correlation with
  ambulatory pressure.
\newblock {\em Hypertension}, 16(6):692--699, 1990.

\bibitem{simon2015bridging}
S.~Simon, S.~Mittelst{\"a}dt, D.~A. Keim, and M.~Sedlmair.
\newblock Bridging the gap of domain and visualization experts with a liaison.
\newblock In {\em Eurographics Conference on Visualization 2015}, pp. 127--131,
  2015.

\bibitem{smilkov17vis}
D.~Smilkov, S.~Carter, D.~Sculley, F.~B. Vi{\'{e}}gas, and M.~Wattenberg.
\newblock Direct-manipulation visualization of deep networks.
\newblock In {\em International Conference on Machine Learning}, 2016.

\bibitem{strobelt18lstmvis}
H.~Strobelt, S.~Gehrmann, H.~Pfister, and A.~M. Rush.
\newblock Lstmvis: {A} tool for visual analysis of hidden state dynamics in
  recurrent neural networks.
\newblock {\em IEEE Transactions on Visualization and Computer Graphics},
  24(1):667--676, 2018.

\bibitem{suo17health}
Q.~Suo, F.~Ma, G.~Canino, J.~Gao, A.~Zhang, P.~Veltri, and A.~Gnasso.
\newblock A multi-task framework for monitoring health conditions via
  attention-based recurrent neural networks.
\newblock In {\em American Medical Informatics Association Annual Symposium},
  2017.

\bibitem{takeda2010bilirubin}
Y.~Takeda, Y.~Takeda, S.~Tomimoto, T.~Tani, H.~Narita, and G.~Kimura.
\newblock Bilirubin as a prognostic marker in patients with pulmonary arterial
  hypertension.
\newblock {\em BMC Pulmonary Medicine}, 10(1):22, 2010.

\bibitem{tsibouris2014ischemic}
P.~Tsibouris, M.~T. Hendrickse, P.~Mavrogianni, and P.~E. Isaacs.
\newblock Ischemic heart disease, factor predisposing to barrett’s
  adenocarcinoma: A case control study.
\newblock {\em World Journal of Gastrointestinal Pharmacology and
  Therapeutics}, 5(3):183, 2014.

\bibitem{wang18}
F.~Wang, H.~Liu, and J.~Cheng.
\newblock Visualizing deep neural network by alternately image blurring and
  deblurring.
\newblock {\em Neural Networks}, 97:162--172, 2018.

\bibitem{wang16speech}
W.~Wang, S.~Xu, and B.~Xu.
\newblock First step towards end-to-end parametric {TTS} synthesis: Generating
  spectral parameters with neural attention.
\newblock In {\em Interspeech 2016, 17th Annual Conference of the International
  Speech Communication Association, San Francisco, CA, USA, September 8-12,
  2016}, pp. 2243--2247, 2016. doi: {{%
10\hspace{.1pt}\discretionary{.}{%
}{.}\hspace{.4pt}21437\discretionary{/}{%
}{/}Interspeech\hspace{.1pt}\discretionary{.}{%
}{.}\hspace{.4pt}2016\discretionary{%
}{-}{-}134}}


\bibitem{wang17segment}
Y.~Wang, Z.~Luo, and P.-M. Jodoin.
\newblock Interactive deep learning method for segmenting moving objects.
\newblock {\em Pattern Recognition Letters}, 96:66 -- 75, 2017.

\bibitem{wongsuphasawat18model}
K.~Wongsuphasawat, D.~Smilkov, J.~Wexler, J.~Wilson, D.~Man{\'{e}}, D.~Fritz,
  D.~Krishnan, F.~B. Vi{\'{e}}gas, and M.~Wattenberg.
\newblock Visualizing dataflow graphs of deep learning models in tensorflow.
\newblock {\em IEEE Transactions on Visualization and Computer Graphics},
  24(1):1--12, 2018.

\bibitem{xu16vqa}
H.~Xu and K.~Saenko.
\newblock Ask, attend and answer: Exploring question-guided spatial attention
  for visual question answering.
\newblock In B.~Leibe, J.~Matas, N.~Sebe, and M.~Welling, eds., {\em Computer
  Vision -- ECCV 2016}, pp. 451--466. Springer International Publishing, Cham,
  2016.

\bibitem{yan2016learning}
C.~Yan, Y.~Chen, B.~Li, D.~Liebovitz, and B.~Malin.
\newblock Learning clinical workflows to identify subgroups of heart failure
  patients.
\newblock In {\em AMIA Annual Symposium Proceedings}, vol. 2016, p. 1248, 2016.

\bibitem{zintgraf17cnn}
L.~M. Zintgraf, T.~S. Cohen, T.~Adel, and M.~Welling.
\newblock Visualizing deep neural network decisions: Prediction difference
  analysis.
\newblock In {\em International Conference on Learning Representations}, 2017.

\end{thebibliography}

\mbox{}
\clearpage
\newpage

\appendix
\appendixpage
\addappheadtotoc

\section{Model Description}
\label{app:model_description}
This section describes the mechanism of a basic recurrent neural network (RNN). We also provide a larger diagram of our proposed \retainex for better understanding.

\subsection{Recurrent neural networks}
\label{app:rnn}

RNNs have been used in numerous prediction tasks of different domains
that require processing sequential data. One of its characteristics is that it can take in sequential vectors as input, producing one or multiple vector-length outputs. Combined with nonlinear functions such as softmax or sigmoid, RNNs can be used to model both classification and regression problems. 

A typical RNN model takes in a sequence of $m-$dimensional
vectors $\mathbf{v}_{1},\mathbf{v}_{2},\ldots,\mathbf{v}_{T}$ consisting of real numbers in chronic or sequential order, and for each input returns a corresponding hidden state vector. That is, when inserting the $t-$th vector $\mathbf{v}_{t}$, its corresponding hidden state $\mathbf{h}_{t}$ is obtained using matrix-vector multiplication as 
\begin{equation}
\mathbf{h}_{t}=\mathrm{tanh}\left(\mathbf{W} \mathbf{v}_{t}+\mathbf{U} \mathbf{h}_{t-1}+\mathbf{b} \right),\label{eq:1}
\end{equation}
where $\mathbf{W} \in\mathbb{R^{\mathit{n\times m}}}$, $\mathbf{U} \in\mathbb{R^{\mathit{n\times n}}}$
and $\mathbf{b} \in\mathbb{R^{\mathit{n}}}$ are all learnable
parameters. Note that while the sizes of $m$ and $n$ do not have to be the same size, in many cases they are fixed to same numbers, usually set to the power of 2 such as 64, 128 or 256. These account to the vector size, and thus the representability of the hidden states. However, while an increased hidden state size leads to additional memory consumption, it does not always guarantee performance improvements, and thus is left as a hyperparameter to be tuned by the model designer. In this paper, we use the notation
\begin{equation}
\mathbf{h}_{1},\mathbf{h}_{2},\ldots,\mathbf{h}_{T}=\mathrm{RNN}\left(\mathbf{v}_{1},\mathbf{v}_{2},\ldots,\mathbf{v}_{T}\right),\label{eq:2}
\end{equation}
to formulate an RNN module that takes in $T$ inputs and creates $T$ hidden states.

For binary prediction tasks as in our proposed work, an additional parameter
$\mathbf{w}_{out}\in\mathbb{R^{\mathit{n}}}$ is
applied to either the last hidden state or the sum of all hidden
states for inner product multiplication. The result is a scalar value, on which we apply the sigmoid
function to obtain a continuous value between 0 and 1. Comparing our obtained value with the correct answer, a real number which is either 0 or 1. we calculate the cross-entropy loss between our obtained value and the correct answer to obtain the loss of our prediction as
\begin{equation}
\boldsymbol{\mathfrak{\mathcal{\mathscr{L}}}}=-\frac{1}{N}\sum_{i=1}^{N}y_{i}\log\left(\hat{y_{i}}\right)+(1-y_{i})\log(1-\hat{y_{i}})
\label{eq:3}
\end{equation}
Using this loss, we obtain the partial gradients with respect to all learnable parameters using gradient descent. After the gradients of the parameters denoted as $\mathbf{W}$ are derived, an update value for each parameter, which is the gradient of a parameter multiplied by a fixed learning rate value, is obtained. This update value is added to the original parameter, which corresponds to the process of `updating a parameter'. By repeating this process of parameter updates using different data samples, the loss of our RNN model consistently decreases and the performance of our model is improved. We describe this process as `training an RNN model'.

Variants of RNNs such as long-short term memory\,(LSTM)\,\cite{hochreiter1997lstm} and
gated recurrent units\,(GRU)\,\cite{cho2014gru} have shown to have the same characteristics as the original model, while ensuring better performance. In this paper, we use the term RNN comprehensively to include these variants. We also mention that while our \retainex model is built on a GRU, we still call it an RNN-based model throughout our paper.

\subsection{Computational backgrounds of bidirectional RNNs}
\label{app:bidirectionality}
An increasing trend in RNNs is the shift towards bidirectional models.
Compared to traditional RNNs which are limited to processing the input sequence in one direction (i.e., from the first input to the last input), bidirectional RNNs\,(biRNNs) introduce another set of hidden
state vectors that are computed by starting from the last input and processing backwards. This is obtained as
\[
\mathbf{h^{\mathrm{f}}}_{1},\mathbf{h^{\mathrm{f}}}_{2},\ldots,\mathbf{h^{\mathrm{f}}}_{T}=\mathrm{RNN_{f}}\left(\mathbf{v}_{1},\mathbf{v}_{2},\ldots,\mathbf{v}_{T}\right),
\]
\begin{equation}
\mathbf{h^{\mathrm{b}}}_{T},\mathbf{h^{\mathrm{b}}}_{T-1},\ldots,\mathbf{h^{\mathrm{b}}}_{1}=\mathrm{RNN_{b}}\left(\mathbf{v}_{T},\mathbf{v}_{T-1},\ldots,\mathbf{v}_{1}\right),\label{eq:4}
\end{equation}
 where the final hidden state vector $\mathbf{h}_{t}$ for each timestep is obtained
by concatenating the two hidden states $\mathbf{h^{\mathrm{f}}}_{t}$ and $\mathbf{h^{\mathrm{b}}}_{t}$, resulting from the forward
and backward RNNs.

\section{Qualitative results}
\label{app:qualitative}

This section provides two tables that show results of our experiments.


\begin{table}[htb]
\small
\centering

\begin{tabular}{|>{\centering}m{.2\linewidth}|>{\centering}m{.4\linewidth}|>{\centering}m{.2\linewidth}|}
\hline 
 Code Type & Code Name & Mean Score\tabularnewline
\hline 
\multirow{5}{*}{Diagnosis} & Hypertensive diseases  & 0.532\tabularnewline
\cline{2-3} 
 & \shortstack{Diseases of oesophagus,\\ stomach and duodenum} & 0.218\tabularnewline
\cline{2-3} 
 & Ischaemic heart diseases & 0.200\tabularnewline
\cline{2-3} 
 & Metabolic disorders & 0.118\tabularnewline
\cline{2-3} 
 & Cerebrovascular diseases & 0.091\tabularnewline
\hline 
\multirow{5}{*}{Treatment} & \shortstack{Outpatient care -\\ established patient} & 0.286\tabularnewline
\cline{2-3} 
 & Glucose (Quantitative) & 0.097\tabularnewline
\cline{2-3} 
 & Hematocrit  & 0.094\tabularnewline
\cline{2-3} 
 & Prothrombin Time  & 0.088\tabularnewline
\cline{2-3} 
 & \shortstack{Electrolyte examination\\ (Phosphorus)} & 0.076\tabularnewline
\hline 
\multirow{5}{*}{Prescription} & Bisoprolol hemifumarate & 0.156\tabularnewline
\cline{2-3} 
 & Aspirin (enteric coated) & 0.081\tabularnewline
\cline{2-3} 
 & Atorvastatin (calcium) & 0.059\tabularnewline
\cline{2-3} 
 & Carvedilol & 0.047\tabularnewline
\cline{2-3} 
 & Rebamipide & 0.032\tabularnewline
\hline 
\end{tabular}
\caption{Top-5 contribution scores averaged over the total number of \underline{patients}.}
\label{tab:top5featsscores_per_person}

\end{table}

\begin{table}[htb]
\small
\centering
\begin{tabular}{|>{\centering}m{.2\linewidth}|>{\centering}m{.4\linewidth}|>{\centering}m{.2\linewidth}|}
\hline 
 Code Type & Code Name & Mean Score\tabularnewline
\hline 
\multirow{5}{*}{Diagnosis} & \shortstack{Obesity and other\\ hyperalimentation} & 0.206\tabularnewline
\cline{2-3} 
 & Other infectious diseases & 0.169\tabularnewline
\cline{2-3} 
 & Ischaemic heart diseases & 0.156\tabularnewline
\cline{2-3} 
 & Hypertensive diseases & 0.134\tabularnewline
\cline{2-3} 
 & Disorders of thyroid gland & 0.119\tabularnewline
\hline 
\multirow{5}{*}{Treatment} & Prothrombin Time & 0.299\tabularnewline
\cline{2-3} 
 & \shortstack{24hr blood pressure\\ examination} & 0.278 \tabularnewline
\cline{2-3} 
 & CA-19-9 & 0.253\tabularnewline
\cline{2-3} 
 & CK-MB & 0.198\tabularnewline
\cline{2-3} 
 & \shortstack{Fibrinogen examination\\ (functional)} & 0.185\tabularnewline
\hline 
\multirow{5}{*}{Prescription} & Bisoprolol hemifumarate & 0.523\tabularnewline
\cline{2-3} 
 & Isosorbide mononitrate & 0.243\tabularnewline
\cline{2-3} 
 & Amlodipine besylate & 0.210\tabularnewline
\cline{2-3} 
 & Mmorphine sulfate & 0.164\tabularnewline
\cline{2-3} 
 & Carvedilol & 0.157\tabularnewline
\hline 
\end{tabular}
\caption{Top-5 contribution scores averaged over the total number of \underline{occurrences}.}
\label{tab:top5featsscores_per_app}
\end{table}

\subsection{Large-scale diagram of \retainex}
\label{app:model_desc_retain}
\begin{figure*}[tb]
\centering
\begin{annotatedFigure}
	{\includegraphics[width=1\linewidth, trim={0cm 0cm 0cm 0cm},clip]{model_new3.pdf}}
	\annotatedFigureBox{0.357,0.4192}{0.6145,0.6092}{A}{0.357,0.6092}
	\annotatedFigureBox{0.116,0.4977}{0.196,0.6077}{B}{0.116,0.6077}
	\annotatedFigureBox{0.644,0.5193}{0.684,0.6093}{C}{0.644,0.6093}
	\annotatedFigureBox{0.362,0.7797}{0.562,0.8997}{D}{0.362,0.8997}
\end{annotatedFigure}
\caption{Overview of \retainex. (A) Using separate embedding matrices,
the binary vectors $\mathbf{x}_{1},\ldots,\mathbf{x}_{T}$ are transformed
into embedding vectors $\mathbf{v^{\mathit{a}}}_{1},\ldots,\mathbf{v^{\mathit{a}}}_{T}$
and $\mathbf{v^{\mathit{b}}}_{1},\ldots,\mathbf{v^{\mathit{b}}}_{T}$,
with time interval information appended to the former. (B) $\mathbf{v^{\mathit{a}}}_{1},\ldots,\mathbf{v^{\mathit{a}}}_{T}$
are fed into a bidirectional RNN to produce scalar weights $\alpha$.
(C) $\mathbf{v^{\mathit{a}}}_{1},\ldots,\mathbf{v^{\mathit{a}}}_{T}$
are fed into another biRNN, this time to generate vector weights $\boldsymbol{\beta}$.
(D) $\alpha$, $\boldsymbol{\beta}$ and $\mathbf{v^{\mathit{b}}}$
are multiplied over all timesteps, then are summed to form a single
vector $\mathbf{o}$, which goes through linear and nonlinear
transformation to produce a probability score $\hat{y}$. }
\label{app_fig:overview}
\end{figure*}

We provide the model architecture in a large-scale diagram in the next page.

\newpage

\end{document}